\documentclass[10pt,journal,compsoc]{IEEEtran}
\usepackage{booktabs}
\usepackage{ifpdf}
\ifCLASSINFOpdf
\else
\fi

\usepackage{adjustbox}
\usepackage{amsmath}
\usepackage{amsopn}
\usepackage{amssymb}
\usepackage{amsthm}
\usepackage{bm}
\usepackage[justification=justified]{caption}
\usepackage{cite}
\usepackage{color}
\usepackage{enumerate}
\usepackage{footnote}
\usepackage{graphicx}
\usepackage{hyperref}
\usepackage{latexsym}
\usepackage{multirow}
\usepackage{subfigure}
\usepackage{tabulary}
\usepackage{threeparttable}
\usepackage{url}
\usepackage[normalem]{ulem}
\usepackage{verbatim}

\newcommand{\tabincell}[2]{\begin{tabular}{@{}#1@{}}#2\end{tabular}} 

\usepackage{xspace}
\usepackage{enumitem}
\makeatletter
\DeclareRobustCommand\onedot{\futurelet\@let@token\@onedot}
\def\@onedot{\ifx\@let@token.\else.\null\fi\xspace}
\def\eg{\emph{e.g}\onedot} 
\def\ie{\emph{i.e}\onedot} 
 
\def\etc{\emph{etc}\onedot} 
 
\def\etal{\emph{et al}\onedot}
\makeatother

\begin{document}
	
	\title{Towards a Robust Deep Neural Network in Texts: A Survey}
	
	\author{Wenqi~Wang\IEEEauthorrefmark{2}\IEEEauthorrefmark{4},
		Run~Wang\IEEEauthorrefmark{2}\IEEEauthorrefmark{4}\IEEEauthorrefmark{1},
		Lina~Wang\IEEEauthorrefmark{2}\IEEEauthorrefmark{4}\IEEEauthorrefmark{1},~\IEEEmembership{Member,~IEEE,}
		Zhibo~Wang\IEEEauthorrefmark{4},~\IEEEmembership{Member,~IEEE,}
		Aoshuang~Ye\IEEEauthorrefmark{2}\IEEEauthorrefmark{4}
		\thanks{\IEEEauthorrefmark{2} W. Wang, R. Wang, L. Wang, and A. Ye are with Key Laboratory of Aerospace Information Security and Trusted Computing, Ministry of Education}
		\thanks{\IEEEauthorrefmark{4} W. Wang, R. Wang, L. Wang, Z. Wang, and A. Ye are with School of Cyber Science and Engineering, Wuhan University, China. E-mail: \{wangwenqi\_001, wangrun, lnwang, zbwang, yasfrost\}@whu.edu.cn}
		\thanks{\IEEEauthorrefmark{1} Run Wang and Lina Wang are the corresponding authors.}}
	
	\maketitle
	
	\begin{abstract}
		
		Deep neural networks (DNNs) have achieved remarkable success in various tasks (\eg{}, image classification, speech recognition, and natural language processing (NLP)). However, researchers have demonstrated that DNN-based models are vulnerable to adversarial examples, which cause erroneous predictions by adding imperceptible perturbations into legitimate inputs. Recently, studies have revealed adversarial examples in the text domain, which could effectively evade various DNN-based text analyzers and further bring the threats of the proliferation of disinformation. In this paper, we give a comprehensive survey on the existing studies of adversarial techniques for generating adversarial texts written by both English and Chinese characters and the corresponding defense methods. More importantly, we hope that our work could inspire future studies to develop more robust DNN-based text analyzers against known and unknown adversarial techniques. 
		
		We classify the existing adversarial techniques for crafting adversarial texts based on the perturbation units, helping to better understand the generation of adversarial texts and build robust models for defense. In presenting the taxonomy of adversarial attacks and defenses in the text domain, we introduce the adversarial techniques from the perspective of different NLP tasks. Finally, we discuss the existing challenges of adversarial attacks and defenses in texts and present the future research directions in this emerging and challenging field.
		
	\end{abstract}
	
	
	\IEEEpeerreviewmaketitle
	
	\section{Introduction} \label{introduction}
	
\begin{figure}[t]
		\centering
		\subfigure[original input]{
			\begin{minipage}[t]{0.5\linewidth}
				\centering
				\includegraphics[width=\linewidth]{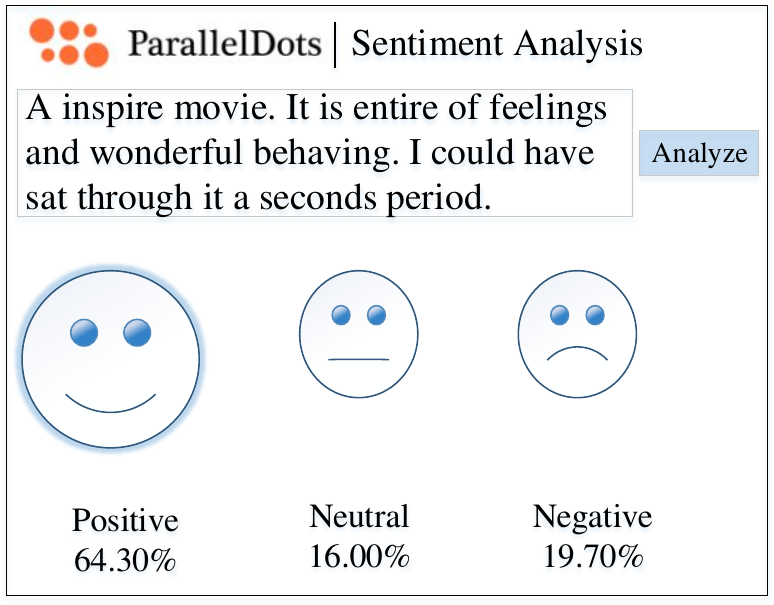}
			\end{minipage}%
		}%
		\subfigure[adversarial text]{
			\begin{minipage}[t]{0.5\linewidth}
				\centering
				\includegraphics[width=\linewidth]{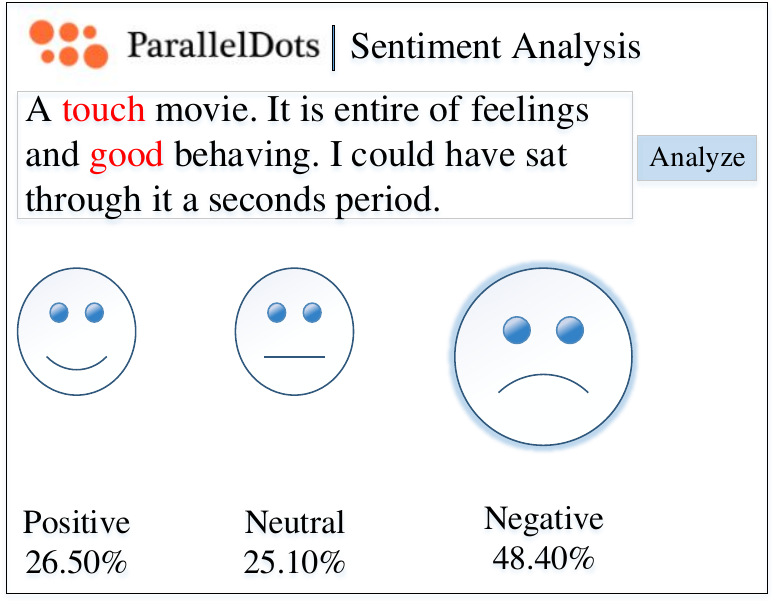}
			\end{minipage}%
		}%
		\centering
		\caption{Instance of an adversarial attack on the popular text analysis system, ParallelDots. ParallelDots provides a series of APIs for various NLP tasks (\eg{}, sentiment analysis) that have achieved state-of-the-art (SOTA) performance. We employ a popular adversarial technique based on the genetic algorithm \cite{maysaebkmskwc2018} to craft adversarial texts and evade the ParallelDots. We can find that the text is predicted as negative in high confidence when the words \emph{inspire} and \emph{wonderful} in the original input are simply replaced by \emph{touch} and \emph{good}, respectively.}
		\label{isnatncefigure}
	\end{figure}

	Nowadays, deep neural networks (DNNs) have shown their great power in addressing masses of challenging problems in various areas, such as computer vision \cite{akisgeh2012,srkhrgjs2015}, audio \cite{ghlddygedarm2012,avdosdhzks2016}, and natural language processing (NLP) \cite{isovqvl2014,hxmddzakaic2016}. Due to their tremendous success, DNN-based systems are widely deployed in the physical world, including many security-critical areas \cite{dctxlsjmxw2006,ezpcbbdsfr2015,jskBerlin2015,zyylzwyx2014,kdjjljsbmpomjk2016}. However, a series of studies \cite{czwzisjbd2014,ijgjscz2015} have found that crafted inputs by adding imperceptible perturbations could easily fool DNNs. These modified inputs are so-called adversarial examples, which bring potential security threats to DNN-based systems even in the black-box scenario where the target system is not available to attackers. For example, Figure \ref{isnatncefigure} shows an adversarial attack on the physical sentiment analysis system named ParallelDots\footnote{\url{https://www.paralleldots.com}}. In this case, we cannot obtain any knowledge of the system architecture, model parameters, and training data. However, it fails to distinguish the adversarial example correctly and output erroneous results. In fighting against the threats of adversarial examples, researchers have conducted numerous works on attacks and defenses, leading to a dramatic increase in both theory and application techniques, varying from images to texts. Here, we focus on the adversarial examples in the text domain rather than the well-investigated image domain. 
	
	In NLP, DNNs are widely employed in many fundamental tasks (\eg{}, text classification, natural language inference, and machine translation). Unfortunately, these DNN-based systems suffer obvious performance degradation in facing adversarial examples. Papernot \etal{} \cite{nppmas2016} first found that attackers could generate adversarial examples by adding imperceptible noises into texts, which would induce classifiers to produce incorrect results. Then, an arms race starts in the text domain battleground, resulting in the exposure of studies in this emerging field. Most of the adversarial attacks in texts focus on specific NLP tasks \cite{jgjlmlsyjq2018,rjpl2017,pmtdtrsr2017,nivenprobing2019}, which will bring potential security concerns to our users. For instance, in the real world, when booking food online, users tend to search for nearby recommended restaurants in mobile apps and read reviews of their products. The service providers \cite{xhzhjszlyjjc2018,swjjtywhl2018,dlygfznjyxxtzyr2018} will give suggestions according to the posted comments via various techniques like sentiment analysis \cite{wmahhk2014}. However, these DNN-based text analyzers could be easily fooled by adversarial examples. Attackers can interfere with product ratings by posting adversarial texts. More seriously, attackers can maliciously propagate disinformation via adversarial texts to reap profits and cause profit losses to consumers. Thus, effective defense methods need to be devised, and robust models should be developed for the community.
	
\begin{figure}[t]
		\centering
		\subfigure[publications in all areas]{
			\begin{minipage}[t]{0.5\linewidth}
				\centering
				\includegraphics[width=\linewidth]{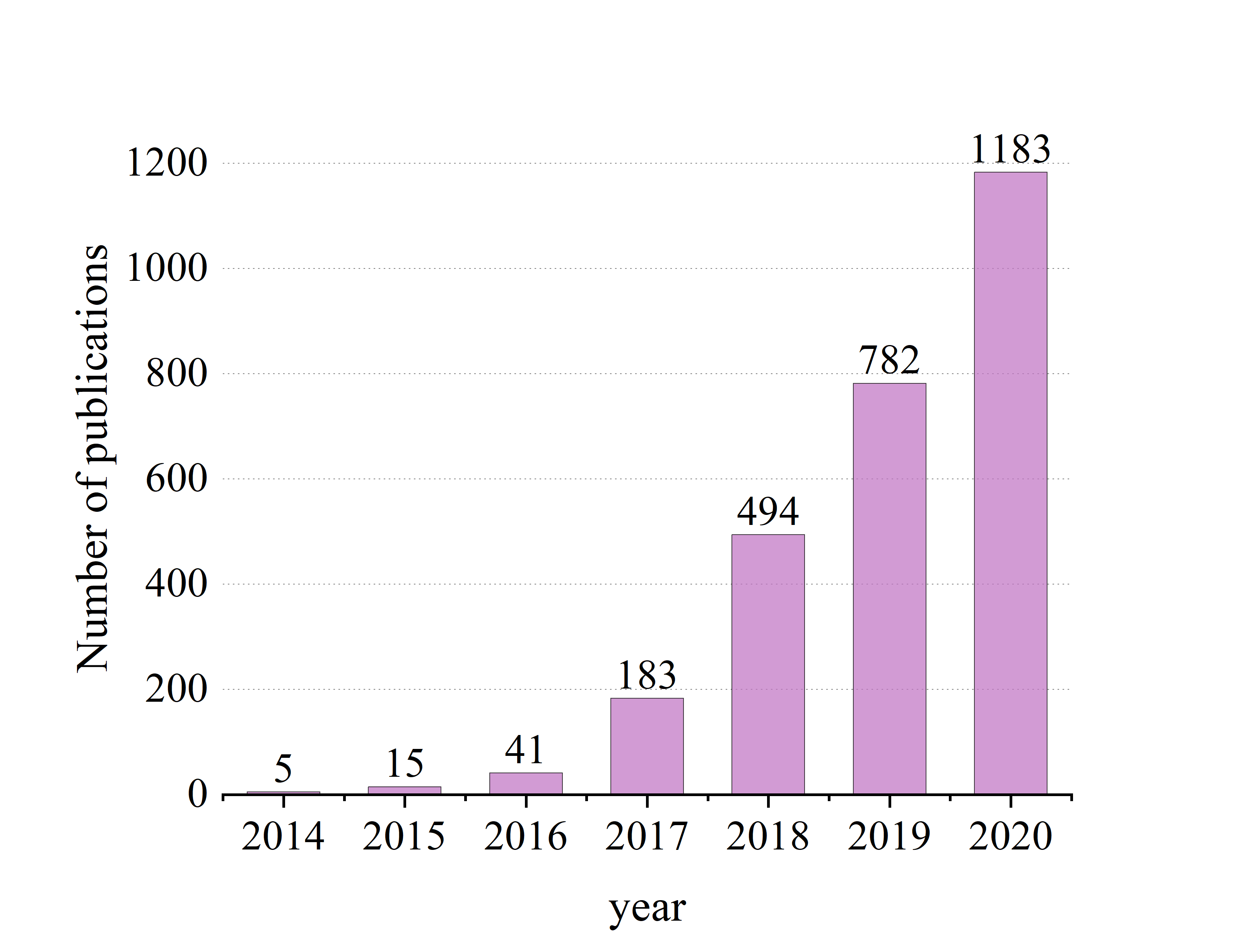}
			\end{minipage}%
		}%
		\subfigure[publications in texts]{
			\begin{minipage}[t]{0.5\linewidth}
				\centering
				\includegraphics[width=\linewidth]{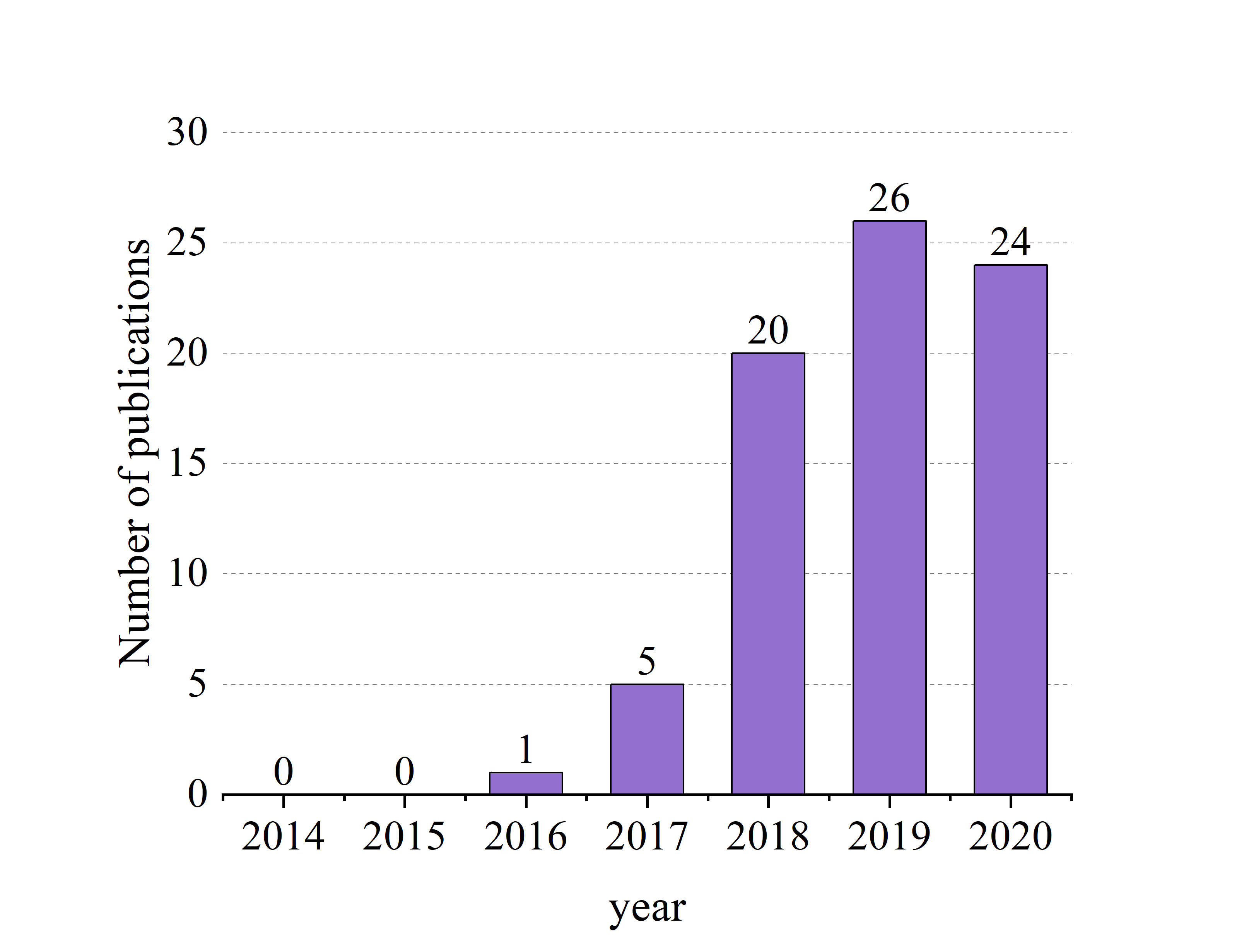}
			\end{minipage}%
		}%
		\centering
		\caption{Publications of adversarial examples. Figure \ref{statisticsjpg}(a) shows the number of publications in the field of adversarial example, which is collected by Carlini\cite{NicholasCarlini2019}, covering a wide range such as image, audio, text, \etc{}. Figure \ref{statisticsjpg}(b) represents the number of publications in the adversarial text domain.}
		\label{statisticsjpg}
	\end{figure}

	For defense, countermeasures have been proposed to enhance the robustness of DNN-based text analyzers. Nevertheless, they are obviously not prepared for the emerging threats of adversarial examples, so that continuous efforts should be taken further. Figure \ref{statisticsjpg} shows us the publications of adversarial examples in recent years, and it reveals that numerous studies are developing various adversarial techniques which pose challenges to defense. At present, adversarial texts detection \cite{dpbdzcl2019} and model enhancement \cite{ijgjscz2015} are two mainstream ideas in fighting against the threats of adversarial texts, but both of them exhibit obvious weakness. For instance, adversarial text detection is only suitable for certain adversarial attacks. Model enhancement like adversarial training suffers the shortcoming in distinguishing adversarial texts generated by unknown adversarial techniques. In summary, tackling unknown adversarial techniques, generalized to different languages, and effective to a wide range of NLP tasks are the three obstacles for the existing defense methods. To bridge this striking gap, it is urgent to inspire researchers to invest in the study of adversarial attacks and defenses in the text domain. Thus, a comprehensive survey is needed to present the preliminary knowledge and introduce the challenges of this field.
	
	
	In adversarial attacks and defenses, several surveys focus on the image domain \cite{naam2018,bbfr2018,jgrpaigdaged2018,qlplwzwcsyvl2018,xyyphqzxl2019,jlzxxj2019}, but few in texts \cite{ybjglass2019,hxymhlddhljtaj2019,wezqzsaacl2019}. Here, we introduce these three surveys in texts and list the differences between them. 
	\begin{itemize}
		\item In 03/2019, Belinkov \etal{} \cite{ybjglass2019} mainly focused on the interpretability of machine learning in NLP. They only review some attacks to understanding these models' failure, but their work lacks surveying the defense methods against adversarial attacks.
		\item In 03/2020, Xu \etal{} \cite{hxymhlddhljtaj2019} systematically reviewed cutting-edge algorithms in the field of images, graphics, and texts. For adversarial attacks in texts, they only describe some methods according to different NLP tasks, but they do not analyze which kind of attack is suitable for the task, nor do they compare the similarities and differences between these methods. Meanwhile, the authors also do not pay attention to the defense in the text domain.
		\item In 04/2020, Zhang \etal{} \cite{wezqzsaacl2019} mainly compared attack methods in the image domain and described how adversarial attacks were implemented in texts. They divide adversarial attacks into black-box and white-box attacks, just like in the image domain. However, this classification method does not reflect how to generate adversarial examples in NLP. Due to the difference between texts and images, adversarial examples can be classified as char-level, word-level, sentence-level, and multi-level attacks according to the perturbation units in texts. Besides, the specially designed defense method (\ie{}, spelling-check) in NLP is not introduced in their \textit{defense} section.
		\item In addition, all of them lack some important guidelines such as the difference between Chinese-based and English-based adversarial examples, interpretability of adversarial examples, and combination with other interesting works (\eg{}, adding adversarial perturbations into deepfake texts to fool deepfake detectors \cite{wzdtzxrwndmzjwjy2020,zrharhbyfarfcy2019,ywwyfmjxbzqdjg2020}).
	\end{itemize}
	
	In this paper, we review the studies of adversarial examples in the text domain with the goal to build robust DNN-based text analyzers by understanding the generation of adversarial texts, the weakness and strengths of existing defense methods, and the adversarial techniques for different NLP tasks. The advances of our work are summarized as follows.  
	\begin{itemize}
		\item We review not only adversarial attacks and defenses in the text domain, but also interpretation, imperceptibility, and certification works. Our systematic and comprehensive review helps newcomers to understand this research filed.
		\item The prior three surveys only focus on works related to English-based models, and neither of them reviews the efforts of evaluating the robustness of Chinese-based models. We bridge this gap and analyze the differences of adversarial examples between English-based and Chinese-based models.
		\item We classify the adversarial texts into \textit{char-level}, \textit{word-level}, \textit{sentence-level}, and \textit{multi-level} according to the perturbation units in generating adversarial texts. Additionally, we focus on the adversarial attacks for the different NLP tasks. We hope this could inspire future researchers to understand the generation of adversarial texts and further develop general and effective defense methods for these NLP tasks.
		\item We combine adversarial examples with model analysis methods to study the adversarial attacks and defenses. We review related analysis methods to explore NLP models' behaviors, contributing to proving the rationality of adversarial attack and defense methods.
	\end{itemize}
	
	The rest of this paper is organized as follows. We first give the preliminary knowledge of adversarial examples in Section \ref{background}. Section \ref{adversarialattacksintext} reviews the adversarial attacks for text classification. Attacks on other NLP tasks are presented in Section \ref{adversarialexamplesonothertasks}. We introduce the defense methods in Section \ref{defense}. Section \ref{Chinesebasedmodels} shows related works on Chinese-based models, and we analyze the differences from English-based ones. Finally, Section \ref{discussion} discusses our findings and the challenges from the reviewed works, which can shed new light on the following research direction. Section \ref{conclusion} gives a conclusion about our comprehensive survey. 
	
	\section{Preliminaries} \label{background}
	
	This section gives the basic preliminary knowledge of adversarial examples, including the formula descriptions, interpretation and classification of adversarial examples, metrics for measuring the imperceptibility of adversarial texts, and available public datasets in texts. 
	
	\subsection{Formula Descriptions} \label{adversarialexampleformulation}
	
	To present a more intuitive understanding of the definitions, we give some formula descriptions about DNN, adversarial examples, and robustness of DNN models.
	
	\textbf{DNN.} A typical DNN can be presented as the function $F: X \rightarrow Y$, which maps from an input set $X$ to a label set $Y$. $\emph{Y}$ is a set of $k$ classes like $\{1, 2, \ldots, k\}$. For a sample $\emph{x} \in X$, it is correctly classified by $\emph{F}$ to the truth label $\emph{y}$, \ie{}, $F(x)=y$. 
	
	\textbf{Adversarial Examples.} An attacker aims at adding the small perturbation $\varepsilon$ in $\emph{x}$ to create adversarial example $\emph{x'}$, such that $F(\emph{x'}) = \emph{y'}(\emph{y} \not= \emph{y'})$. At the same time, $\emph{x'}$ not only needs to fool $F$ but also should be imperceptible to humans. To enforce the generated $\emph{x'}$ to be imperceptible, a series of metrics (\eg, semantic similarity) are adopted to achieve this goal, \ie{}, $\parallel\varepsilon\parallel < \delta$. $\delta$ is a threshold to limit the size of perturbations.

	\textbf{Robustness.} In defending against adversarial examples, a robust DNN model should tolerate the adversarial attacks and output the correct predictions in tackling the imperceptible additive noises \cite{DouglasHeaven2019}. Hence, the prediction of the adversarial example $\emph{x'}$ should be $y$ rather than $y'$ in a robust DNN model, \ie{}, $F(x')=y$. The defense methods to enhance the robustness of models should tackle a wide range of $\varepsilon$ effectively.
	
	\subsection{Interpretation of Adversarial Examples}
	
	Answering why adversarial examples exist can help us devise more effective and practical defense methods. In recent years, researchers have been continuously exploring this question since they observed the adversarial examples in 2014. However, the existence of adversarial examples is still an open question to the community. Here, we briefly introduce the recent efforts in exploring this question.
	
	\begin{enumerate}[leftmargin=*]
		\item \textbf{Model Linear Hypothesis.} Goodfellow \etal{}\cite{ijgjscz2015} proposed the linearity hypothesis and claimed that the existence of adversarial examples was the linear behavior of DNNs in high-dimensional space. The classifier is not sensitive to adversarial perturbations added to each input dimension, but it will misbehave when the perturbations are applied to all dimensions. Other studies like \cite{8850530,amsGanguli2017} also support the linear hypothesis contributing to the vulnerability of DNNs. However, Sabour \etal{} \cite{ssycffdjf2016} doubled this point and demonstrated that the linearity hypothesis did not apply to their work because of the internal representation of adversarial examples in the DNN.
		
		\item \textbf{Data Distribution Influence.} Shafahi \etal{} \cite{aswrhcssftg2019} claimed that DNN models were vulnerable to adversarial examples due to the data distribution. For a dataset, if adjacent pixels in images are highly correlated, the model is relatively robust against adversarial examples based on these images. While the pixels spread out and have less correlation, the vulnerability to adversarial examples will increase. However, this could explain the existence of adversarial examples in images well, but it is not applicable to texts.
		
		\item \textbf{Input Features.} Ilyas \etal{} \cite{aissdilebtam2019} demonstrated that adversarial examples were not bugs, but features. The features can be classified as robust and fragile in prediction. Both of them could be used for prediction, but adversarial examples are generated when the perturbations are added to the fragile features. Thus, the adversarial examples widely exist in image, text, or other domains.
	\end{enumerate}
	
	However, the reason for the existence of adversarial examples is still not clear, though continuous efforts have been paid in recent years. Recently, the DNN models have achieved tremendous success in many challenging tasks, but it is still a black-box to us, which draws continuous efforts to open it. Experience simply tells us that the deeper and wider network is the master key for improving performance. Thus, it is difficult to understand why our DNN models are susceptible to adversarial examples. Understanding the working mechanism of DNNs will be a key step in identifying the existence of adversarial examples. We hope that our survey could inspire future researchers to investigate this open question, which will promote the usage of DNNs in safety-critical areas.
	
	\subsection{Transferability of Adversarial Examples}
	
	Szegedy \etal{}\cite{czwzisjbd2014} first found that adversarial examples generated from a neural network could also make another network misbehave by different datasets, reflecting their transferability. Therefore, attackers can train a substitute model and utilize the transferability of adversarial examples for the attack when they have no access and query restriction to target models. Recently, studies show that different types of adversarial attacks have different transferability \cite{akigsb2017,lwzxzctwe2018}. For instance, adversarial examples generated from one-step gradient-based methods are more transferable than iterative methods \cite{akigsb2017}, but their attack abilities are the opposite. Hence, the generation of adversarial examples with high transferability is not only the premise to carry out black-box attacks, but also a metric to evaluate generalized attacks.
	
	\subsection{Taxonomy of Adversarial Examples}
	
\begin{figure}[t]
		\centering
		\setlength{\belowcaptionskip}{-0.3cm}
		\includegraphics[width=\linewidth]{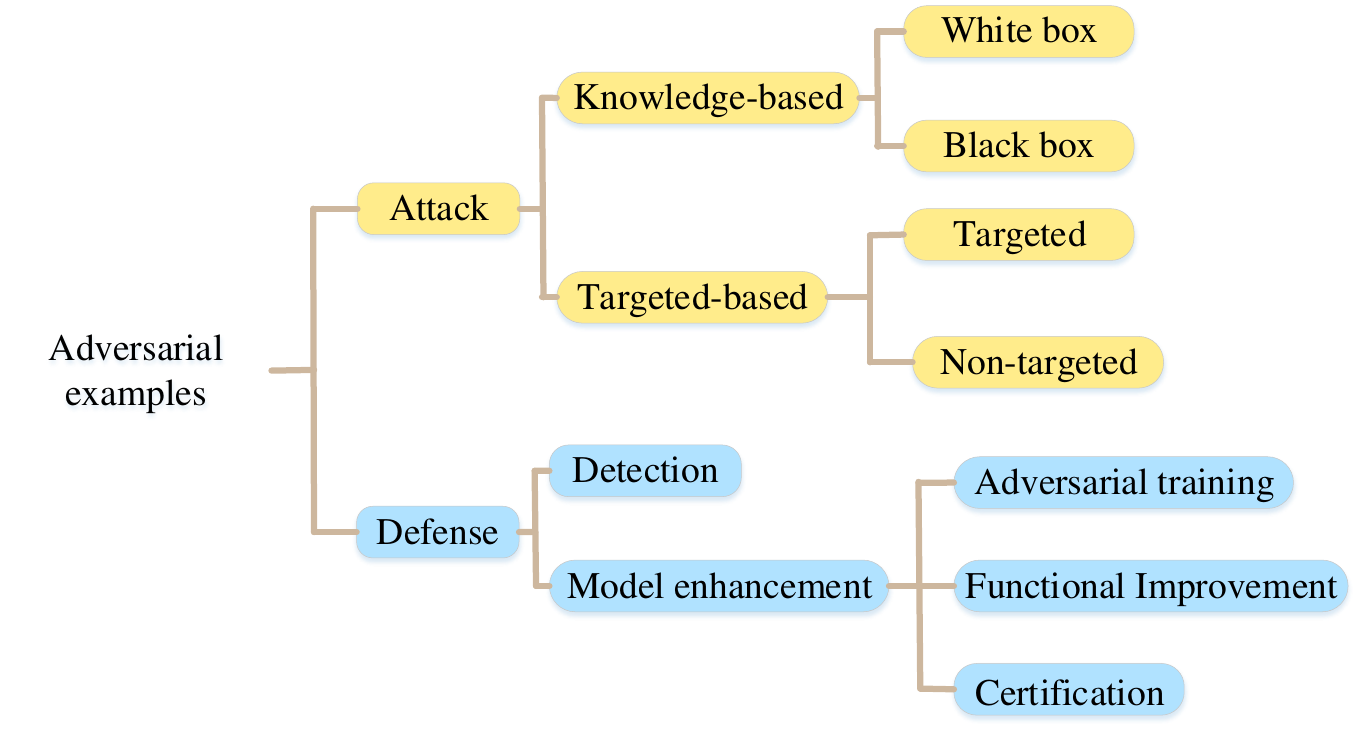}
		\caption{Classification of adversarial attack and defenses.}	\label{fig2}
	\end{figure}

	Figure \ref{fig2} presents a classification of adversarial attacks and defenses. The adversarial attacks could be easily divided into black-box and white-box attacks according to the knowledge of target models. Additionally, the attacks could also be divided into targeted attacks and non-targeted attacks based on whether the erroneous output is desired. In defending against adversarial examples, the mainstream ideas are adversarial example detection and model enhancement. Adversarial example detection indicates observing the minor difference between legitimate inputs and adversarial examples. Model enhancement denotes modifying the model architecture and updating the parameters in a model.
	
	\subsubsection{Taxonomy of Adversarial Attacks} \label{typesofadversarialattack}
	
	Adversarial attacks can be conducted in both white-box and black-box scenarios. In the white-box scenario, adversaries have full access to target models. They can generate perfect adversarial examples by leveraging target models' knowledge, including model architectures, parameters, and training data. In the black-box scenario, adversaries can not obtain any knowledge of the target models. They utilize the transferability \cite{ijgjscz2015} of adversarial examples or repeated queries for optimization to perform a black-box attack. 
	
	According to the desire of adversaries, adversarial attacks can be divided into targeted and non-targeted attacks. In the targeted attack, the generated adversarial example $\emph{x'}$ is purposefully classified into a specified class $t$, which is the adversary's target. This process mainly relies on increasing the confidence score of class $t$. In the non-targeted attack, the adversary only aims at fooling the model rather than expect the desired output. The result $\emph{y'}$ can be any class except for $\emph{y}$. Contrary to the targeted attack, the non-targeted attack operates via reducing the confidence score of the correct class $y$.
	
\begin{figure}[t]
		\centering
		\setlength{\belowcaptionskip}{-0.3cm}
		\includegraphics[width=\linewidth]{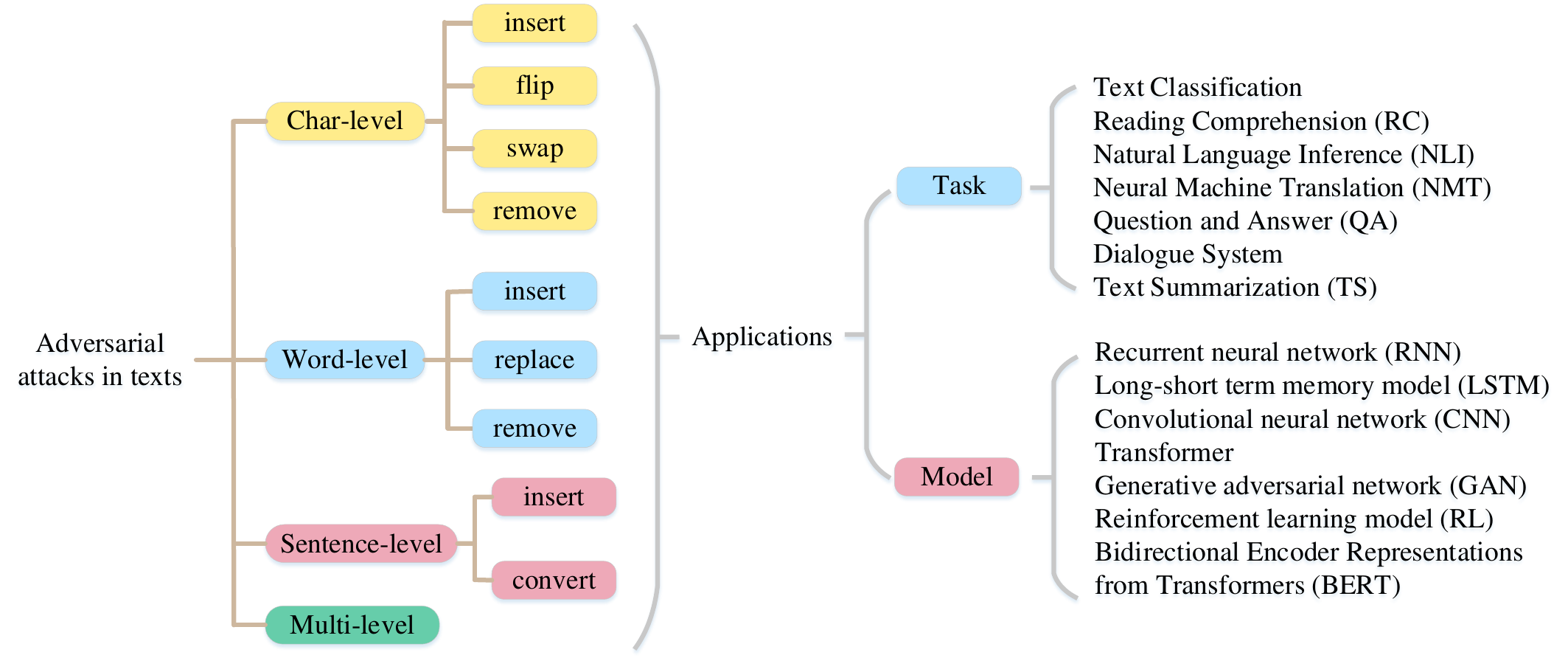}
		\caption{Classification of adversarial texts based on the perturbations units.} \label{AEintexts}
	\end{figure}

	In the text domain, adversarial attacks can be classified as char-level, word-level, sentence-level, and multi-level (shown in Figure \ref{AEintexts}) according to the perturbation units in generating adversarial examples. Char-level attacks indicate that adversaries modify several characters in words to generate adversarial examples that can fool the detectors. Specifically, the modifications are mostly misspellings, and the common operations include insertion, swap, deletion, and flip. Word-level attacks involve various word perturbations. Attackers generate adversarial examples by inserting, replacing, or deleting certain words in various manners. Sentence-level attacks usually insert a sentence into a text or rewrite the sentence while maintaining its meanings. Multi-level attacks incorporate more than one of the three perturbation attacks to achieve the imperceptible and high success rate attack.
	
	\subsubsection{Taxonomy of Defenses against Adversarial Attacks} \label{typesofadversarialdefense}
	
	In defending against adversarial attacks, the goal is to build a robust DNN model in tackling various known and unknown adversarial techniques well \cite{ncaanpwbjrdtigamak2019}. In the existing studies, the mainstream defense strategies could be divided into adversarial example detection and model enhancement. 
	
	Adversarial example detection indicates to directly distinguish adversarial examples from the legitimate inputs based on the observed subtle differences. Model enhancement involves parameters update or architecture modification, such as adversarial training and adding additional layers. In texts, spelling check and adversarial training are two major ways for defending against adversarial attacks. The spelling check is a special detection method in NLP, while adversarial training is a general approach employed in image, text, audio, \etc{}

	\subsection{Metrics on Imperceptibility in texts} \label{metric}
	
	Adversarial examples are crafted by adding imperceptible perturbations into legitimate inputs to incur erroneous output labels \cite{aswrhcssftg2019}. In the image domain, various metrics are adopted to measure the imperceptibility of adversarial examples. $L_{p}$ norm is the most commonly used method defined as
	\begin{equation} \label{triangle}
	\Vert\triangle c \Vert_{p}=\sqrt[p]{\sum_{i=1}^{n} |c'_{i}-c_{i}|^{p}} 
	\end{equation}
	where $\triangle c$ represents the perturbations. $c'_{i}$ and $c_{i}$ are the $i$-th factors in $n$-dimensional vectors $\vec{c'}$ and $\vec{c}$, respectively. Formula \eqref{triangle} represents a series of distances, where $p$ could be 0 \cite{nppmsjmfzbcas2016,jwsdvvsk2017,ncdw2017}, 2 \cite{smmdafofpf2017,ncdw2017,smmdafpf2016,mcyannjk2017}, $\infty$ \cite{czwzisjbd2014,ijgjscz2015,mcyannjk2017,akigsbengio2017}, and so on. Specially, when $p$ is equal to zero, $\Vert\triangle c \Vert_{0}$=$\sum bool(c_{i}\ne 0)$. $bool$ is a logical function with 0 or 1. 
	
	However, it is impossible to borrow the ``imperceptible'' measurements in images to texts. In comparison with the pixel-level modification in images, perturbations for generating adversarial texts operate the characters, words, or sentences, which are visible to humans due to the introduced grammatical and spelling errors. A successful attack needs to maintain the semantic meaning of the generated adversarial texts the same as the original ones and be imperceptible to humans. Therefore, imperceptible adversarial examples (\eg{}, Figure \ref{isnatncefigure}) in the text domain should satisfy the following basic requirements. (1) No obvious errors could be easily observed by human eyes. (2) The crafted adversarial texts should convey the same semantic meaning as the original ones. (3) The model output on the adversarial text and the legitimate input should be different, which means an erroneous output occurred. Thus, the majority of metrics adopted in images can not be directly applied in measuring texts due to the symbolic  representations of perturbations in texts rather than the number representations like the pixel. Next, we detail the metrics (\eg{}, Euclidean distance, Edit distance, Cosine similarity, and Jaccard Similarity Coefficient) employed for measuring the imperceptibility of adversarial texts. 
	
	\textbf{Euclidean Distance}\cite{maysaebkmskwc2018}. The original Euclidean distance is the beeline from one point to another in Euclidean space. As the mapping of text to this space, it acts as a metric to calculate the similarity between two objects, which are represented as vectors. Thus, given two word vectors $\vec{m}=(m_1,\ldots, m_i, \ldots, m_k)$ and $\vec{n}=(n_1,\ldots, n_i, \ldots, n_k)$, the Euclidean distance $ED$ of these two vectors is defined as
	\begin{equation} \label{EuclideanDistance}
	ED\!=\!\sqrt{(m_1\!-\!n_1)^2\!+\!\cdots\!+\!(m_i\!-\!n_i)^2+\!\cdots\!+\!(m_k\!-\!n_k)^2}
	\end{equation}
	where $m_{i}$ and $n_{i}$ are the $i$-th factors in the $k$-dimensional vectors, respectively. The lower the distance is, the more similar they are. 
	
	\textbf{Edit Distance}\cite{jgjlmlsyjq2018}. Edit distance refers to the number of editing operations required to convert one string to another, and Levenshtein distance \cite{vil1966} is a widely used edit distance. For two strings $a$ and $b$, the Levenshtein distance $lev$ is calculated by 
	\begin{eqnarray} \label{editDistance}
	lev(i,j)=\left\{
	\begin{array}{ll}
	\max (i,j),\quad\quad \min (i,j)=0 \\
	\min\left\{ \begin{array}{ll}
	lev(i-1,j)+1 \\
	lev(i,j-1)+1 \quad otherwise. \\
	lev(i-1,j-1)+1_{a_i\neq b_i} 
	\end{array}
	\right.
	\end{array} 
	\right.
	\end{eqnarray}
	where $lev(i,j)$ is the distance between the first $i$ characters in $a$ and the first $j$ characters in $b$. The lower it is, the more similar the two strings are.
	
	\textbf{Cosine Similarity}\cite{djzjjjtzps2020}. Cosine similarity refers to the similarity between two vectors by measuring the cosine of the angle between them. For two given word vectors $\vec{m}$ and $\vec{n}$, the cosine similarity $CD$ is calculated by 
	\begin{equation} \label{cosinesimilarity}
	CD = \frac{\vec{m} \cdot \vec{n}}{\Vert m \Vert \cdot \Vert n \Vert} = \frac{\sum\limits_{i=1}^k m_i \times n_i}{\sqrt{\sum\limits_{i=1}^k (m_i)^2} \times \sqrt{\sum\limits_{i=1}^k (n_i)^2}}   
	\end{equation}
	Compared with Euclidean distance, the cosine similarity pays more attention to the difference between the directions of two vectors. The more consistent their directions are, the more similar they are. 
	
	\textbf{Jaccard Similarity Coefficient}\cite{jfsjtdbltw2019}. The Jaccard similarity coefficient is used to compare the similarity between a limited sample set. For two given sets A and B, their Jaccard similarity coefficient $J(A, B)$ is calculated by
	\begin{equation} \label{JaccardSimilarity}
	J\left(A, B\right) = |A \cap B| / |A \cup B| 
	\end{equation}
	where $0 \leq J(A,B) \leq 1$. The closer the value of $J(A,B)$ is to 1, the more similar they are. In texts, intersection $A \cap B$ refers to similar words in the samples, and union $A \cup B$ is all words without duplication.
	
	These aforementioned metrics are widely applied in tackling various machine learning tasks. Euclidean distance and cosine distance accept vectors for calculation, while the Jaccard similarity coefficient and edit distance directly operate on the raw texts without any transformations into vectors. Particularly, Michel \etal{}\cite{pmxlgnjmp2019} proposed a natural criterion for adversarial texts on sequence-to-sequence models. This work focuses on evaluating the semantic equivalence between adversarial examples and the original ones. Experimental results show that strict constraints are useful for keeping meaning-preserving, but the performance compared with the aforementioned metrics needs further research efforts.
	
	\subsection{Datasets in Texts} \label{datasets}
	
	We survey the top-tier conferences and journals in artificial intelligence (AI) and NLP (\eg{}, ICLR, ACL, AAAI, EMNLP, IJCAI, NAACL, COLING, TACL, TKDE, and JMLR) to collect the employed databases in texts. Table \ref{datasetsinthetable} shows the details of the widely-adopted datasets employed in the studies of adversarial attacks.
	
	\begin{table*}[t]
		\scriptsize
		\centering
		\caption{Twelve popular text datasets employed in the studies of adversarial attacks. The second column shows the name of each data with the source download link. The following three columns give a brief description, size, and application in which NLP task. NLI is short for natural language inference, NMT is short for neural machine translation, and QA is short for question and answer.}
		\label{datasetsinthetable}
		\begin{adjustbox}{width=\linewidth,center}
			\begin{tabular}{|l|l|l|l|l|}
				\hline 
				\multirow{3}{*}{Task} & \multirow{3}{*}{Name} & \multirow{3}{*}{Description} & \multirow{3}{*}{Size}  &  \multirow{3}{*}{Application} \\
				& & & &   \tabularnewline
				& & & &  \tabularnewline
				\hline 
				\hline
				\multirow{8}{*}{classification} & AG's news\footnotemark[2] & News from over 2,000 sources & 144K & \cite{jeardldd2018,Prashanthvdb2019,srydkhwc2019,mbsmmdmsbpfard2019,djzjjjtzps2020,llrmqgxyxxpq2020}  \tabularnewline
				\cline{2-5}
				& DBPedia\footnotemark[2] & Structured content from Wikimedia projects & 45K & \cite{msjshsym2018,blhlmspbxlws2018}  \tabularnewline
				\cline{2-5}
				& Amazon\footnotemark[2] & Product reviews on Amazon & 2 million & \cite{sggRamakrishnan2020}  \tabularnewline
				\cline{2-5}
				& Yahoo\footnotemark[2] & Yahoo! Answers Comprehensive Questions & 1.4 million & \cite{srydkhwc2019,pyjccjhjlmij2020} \tabularnewline
				\cline{2-5}
				& Yelp\footnotemark[2] & User reviews of merchants & 140K & \cite{ylhmcdcjwwwlhcjh2019,djzjjjtzps2020,sggRamakrishnan2020,llrmqgxyxxpq2020,bxwhpbpqcswbl2020} \tabularnewline
				\cline{2-5}
				& IMDB\footnotemark[2] & polarized movie reviews & 50K & \tabincell{l}{\cite{jgjlmlsyjq2018,nppmas2016,sssm2017,msjshsym2018,maysaebkmskwc2018,srydkhwc2019,hzzhznmll2019,djzjjjtzps2020,pyjccjhjlmij2020} \\ \cite{yzfqchyzlmzqlms2020,sggRamakrishnan2020,llrmqgxyxxpq2020,mtrsscg2018,jfsjtdbltw2019,Prashanthvdb2019}} \tabularnewline
				\cline{2-5}
				& MR\footnotemark[3] & movie-review data & 10K & \cite{blhlmspbxlws2018,jfsjtdbltw2019,djzjjjtzps2020,sggRamakrishnan2020,msjshsym2018,mtrsscg2018} \tabularnewline
				\cline{2-5}
				& SST\footnotemark[4] & standard sentiment dataset from Stanford & 240K & \cite{mijwkglz2018,ewsfnkmgss2019,mbsmmdmsbpfard2019,yzfqchyzlmzqlms2020} \tabularnewline
				\hline
				QA & SQuAD\footnotemark[5] & dataset for question answering and reading comprehension from Wikipedia & 100K & \cite{rjpl2017,pkmatmskd2018,ywangmbsan2018,GanandHweeouNg2019,ewsfnkmgss2019,bxwhpbpqcswbl2020,stsjmykrs2020} \tabularnewline
				\hline
				\multirow{2}{*}{NLI}& SNLI\footnotemark[6] & human-written English sentence pairs  & 570K & \cite{zzddssss2018,pmsr2018,mgvsyg2018,maysaebkmskwc2018,ewsfnkmgss2019,hzzhznmll2019,djzjjjtzps2020,yzfqchyzlmzqlms2020,llrmqgxyxxpq2020} \tabularnewline
				\cline{2-5}
				& MultiNLI\footnotemark[7] & crowd-sourced collection of sentence pairs & 433K & \cite{pmsr2018,ylhmcdcjwwwlhcjh2019,djzjjjtzps2020,llrmqgxyxxpq2020} \tabularnewline
				\hline
				NMT & WMT14\footnotemark[8] & parallel texts (\eg{}, German/English) for translation models & -- & \cite{ycljwm2019,wzsjhjxxydjjc2020,stsjmykrs2020} \tabularnewline
				\hline
			\end{tabular}
		\end{adjustbox}
	\end{table*}
	
\begin{figure}[t]
		\centering
		\setlength{\belowcaptionskip}{-0.3cm}  
		\includegraphics[width=\linewidth]{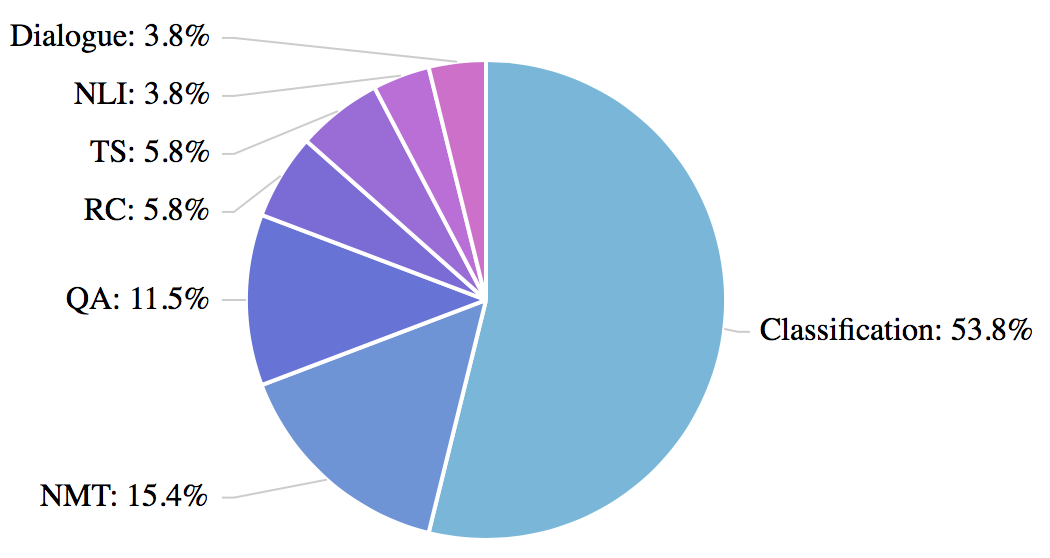}
		\caption{Statistics of datasets used in the research of adversarial attacks. Totally, there are 52 different datasets employed in related works. All of them can be found in Table \ref{tabpaper_information}.} \label{howmanydatasets}
	\end{figure}

	Beyond investigating the popular text databases employed in adversarial attacks, we also explore how many datasets are adopted in the recent works and which tasks they apply. Figure \ref{howmanydatasets} presents the proportion in the corresponding NLP task. We can observe that more than half of the databases focus on text classification, which is a critical NLP task.
	
	\footnotetext[2]{\url{https://course.fast.ai/datasets}}
	\footnotetext[3]{\url{http://www.cs.cornell.edu/people/pabo/movie-review-data/}}
	\footnotetext[4]{\url{https://github.com/stanfordnlp/sentiment-treebank}}
	
	\section{Adversarial Attacks for classification in Texts} \label{adversarialattacksintext}
	
	In recent studies, the majority of adversarial attacks focus on fooling the text classification systems. Next, we detail the four types of adversarial attacks based on the perturbations units: char-level attack, word-level attack, sentence-level attack, and multi-level attack.
	
	\subsection{Char-level Attacks}
	
	Char-level attacks indicate that adversaries modify several characters in words to generate adversarial examples that can fool the detectors. Generally, the modifications are often misspellings, and the operations include insertion, swapping, deletion, and flipping. Although this kind of attack can achieve a high success rate, misspellings can be easily detected. Next, we introduce some representative char-level attacks. 
	
	Gao \etal{}\cite{jgjlmlsyjq2018} proposed a char-level attack, called DeepWordBug, to generate adversarial examples in the black-box scenario, which followed a two-step pipeline. The first stage quantifies the importance of words and determines which one to change. The calculation process for the first stage is shown in 
	\begin{equation} \label{firststage}
	\begin{split}
	CS(x_i)=&[F(x_1,\ldots,x_{i-1},x_i)-F(x_1,x_2,\ldots,x_{i-1})]+\\&\lambda[F(x_i,x_{i+1},\ldots,x_n)-F(x_{i+1},\ldots,x_n)]
	\end{split}
	\end{equation}
	where $CS(x_i)$ represents the importance score of $i$-th word in $(x_{1},\ldots,x_{n})$, evaluated by the function $F$. $\lambda$ is a hyper-parameter. The second stage adds imperceptible perturbations to the selected words through swapping, flipping, deletion, and insertion. Meanwhile, edit distance is used to preserve the readability of generated adversarial examples.  
	
	Gil \etal{}\cite{ygycogjb2019} derived a new method DISTFLIP based on HotFlip \cite{jeardldd2018}. The authors distill the knowledge of the procedure in HotFlip for training their model. Through the trained model, the authors generate adversarial examples to conduct a black-box attack. This method performs better than HotFlip on a toxicity classifier\cite{hhskbzrp2017}, and its run-time in generating adversarial examples is ten times faster than HotFlip. However, the capability to distill the knowledge of any white-box attacks is not clear.
	
	\subsection{Word-level Attacks}
	
	Word-level attacks manipulate the whole word rather than several characters in words. Hence, the modifications are more imperceptible to humans than char-level attacks. The common manipulation includes insertion, deletion, and replacement. According to the way of selecting manipulated words, the word-level adversarial attack can be classified into gradient-based, importance-based, and other attacks.
	
	\footnotetext[5]{\url{https://datarepository.wolframcloud.com/resources/SQuAD-v1.1}}
	\footnotetext[6]{\url{https://nlp.stanford.edu/projects/snli/}}
	\footnotetext[7]{\url{https://cims.nyu.edu/~sbowman/multinli/}}
	\footnotetext[8]{\url{http://www.statmt.org/wmt14/translation-task.html}}
	
	\subsubsection{Gradient-based attacks}
	
	Studies on adversarial examples in the image domain are more active than those in texts. Inspired by the fast gradient sign method (FGSM) \cite{ijgjscz2015} in the image domain, attackers generate adversarial examples by calculating the gradient of text vectors in a model.
	
	As far as we know, Papernot \etal{}\cite{nppmas2016} first studied the problem of adversarial examples in texts and contributed to producing adversarial input sequences. The authors leverage computational graph unfolding\cite{pjw1988} to evaluate the forward derivative \cite{nppmsjmfzbcas2016} (\ie{}, the model’s Jacobian $J(x)$), which is related to the search of modified words. In detail, they utilize FGSM to guide the perturbations, and it can be represented as 
	\begin{eqnarray}
	sign(J(x)[i,\arg\max_{0,1}(p_j)])
	\end{eqnarray} 
	where $i$ is the $i$-th word in an input sequence, and $p_j$ indicates the probability of belonging to category $j$. If the value of $\arg\max_{0,1}(p_j)$ changes, the modification is effective. However, the words in input sequences are iteratively selected for substitution. Hence, there may exist grammatical errors in the generated adversarial examples. 
	
	Different from Papernot \etal{} \cite{nppmas2016}, Samanta \etal{}\cite{sssm2017} employed FGSM to evaluate the important or salient words, which deeply affected the results of classification when they were removed. Three modification strategies (\textit{i.e.}, insertion, replacement, and deletion) are introduced to craft top $k$ words with the highest importance, where $k$ is a threshold. Except for the deletion strategy, both insertion and replacement on top $k$ words require an additional dictionary for operation. Thus, the authors establish a pool of candidates for each word in the experiments, including synonyms, typos, and type-specific keywords. However, the establishment of the candidate pool suffers huge consumption, and there may be no candidate pool for some top $k$ words in the actual inputs.   
	
	Unlike the above methods, Sato \etal{}\cite{msjshsym2018} proposed iAdv-Text by adding perturbations in the embedding space. iAdv-Text formulate this as an optimization problem, which jointly minimizes objection function $\mathcal{J}_{iAdvT}(D,W)$ on the entire training dataset $D$ with parameters $\emph{W}$. The optimization procedure is shown in 
	\begin{equation} \label{eq5}
	\begin{split}
	\mathcal{J}_{iAdvT}(D,W) = &\frac{1}{|D|}\mathop{\arg\min}_{W}\{\sum_{(\hat{X},\hat{Y})\in D}\ell(\hat{X},\hat{Y},W)+\\&\lambda\sum_{(\hat{X},\hat{Y})\in D}\alpha_{iAdvT}\}
	\end{split}
	\end{equation}
	where $\hat{X}$ and $\hat{Y}$ represent the inputs and labels, respectively. $\lambda$ is a hyper-parameter to balance the two loss functions. $\ell(\hat{X},\hat{Y},W)$ is the loss function of individual training sample $(\hat{X},\hat{Y})$ in $D$. $\alpha_{iAdvT}$ is a maximization process to find the worst case weights of the direction vectors calculated by
	\begin{equation} \label{functionofiAdvT}
	\alpha_{iAdvT} = \frac{\epsilon g}{\Vert g \Vert_2}, g = \nabla_{\alpha}\ell(\vec{w} + \sum_{k=1}^{|V|}a_kd_k, \hat{Y}, W)
	\end{equation}
	where $\sum_{k=1}^{|V|}a_kd_k$ is the perturbation generated from each input on its word embedding vector $\vec{w}$, $\epsilon$ is a hyper-parameter to control adversarial perturbations, $a_{k}$ is the $k$-th factor of a $|V|$-dimensional word embedding vector $\alpha$, $d_{k}$ is the $k$-th factor of a $|V|$-dimensional direction vector $\vec{d}$, which is a mapping from one word to another in embedding space. iAdv-Text restricts the direction of perturbations with cosine similarity for finding a substitution, which is in a pre-defined vocabulary rather than an unknown word. 
	
	Behjati \etal{} \cite{mbsmmdmsbpfard2019} designed a universal perturbations added to any input. They optimize the gradient of loss function $loss$ to construct a word vocabulary $V$ containing words that apply to all data. The optimization is shown in 
	\begin{eqnarray}\label{Behjatietal}
	w'_i=\arg\min_{w'_i \in V} cos(emb(w'_i),(emb(w_i)+\alpha r_i))
	\end{eqnarray}
	where $r_i=\nabla_{emb(w_i)}loss(l,f(x'))$. $emb(w_i)$ is the corresponding embedding of word $w_i$, $l$ represents the label. If it is a targeted attack, $l$ is the target label, and the learning rate $\alpha$ is negative. Otherwise, $l$ is the ground truth label, and $\alpha$ is positive. The words in $V$ will insert into texts to generate adversarial examples. However, the insertion occurs at the beginning of the input sequence leading to grammatical errors and breaking the imperceptibility rule.
	
	\subsubsection{Importance-based attacks}
	
	By analyzing the existing methods of generating adversarial texts, researchers noticed that the importance of each word in determining the final predictions is vastly different. Based on this initial idea, researchers have launched successful attacks by computing the importance of words and modifying these high valuable words. Importance-based attacks usually follow a two-step pipeline.
	\begin{enumerate}[leftmargin=*]
		\item Calculating the importance of words by querying the target model multiple times.
		\item Modifying these important words via insertion, deletion, or replacement. 
	\end{enumerate}
	
	Ren \etal{} \cite{srydkhwc2019} designed a synonym replacement method PWWS working on the word-level. The authors construct a synonym set $\mathbb{L}_i$ for each word in the inputs. To search for suitable substitutions, an optimization process in formula \eqref{synonymset} is conducted by maximizing the word saliency of these words.
	\begin{equation} \label{synonymset}
	\begin{split}
	R(w_i,\mathbb{L}_{i})=\arg\max{P(y_{true}|x)-P(y_{true}|x_{i}')}
	\end{split}
	\end{equation}
	where $R(w_i,\mathbb{L}_{i})$ substitutes the best candidate synonym $w_{i}^{*}$ of $i$-th word in the text $x$. $x_{i}'$ is obtained by replacing the $i$-th word in $x$ with each candidate. $P(y|x)$ is the classification probability of $x$. After that, the final process determines the order of replacement in $x$. 
	
	Hsieh \etal{} \cite{ylhmcdcjwwwlhcjh2019} thought that the changes in words with the highest or lowest attention scores could substantially undermine self-attentive models' predictions. Hence, they exploit the attention scores as a potential source of vulnerability and modify target words with the highest or lowest scores. A random word in the vocabulary is greedily selected to replace the target word until the attack succeeded. Although the constraint on the embedding distance is imposed to keep semantically similar, the vocabulary construction is not clear. Similarly, Yang \etal{} \cite{pyjccjhjlmij2020} greedily searched for the weak spot of the input sentence by replacing a word with the padding. If the probability changes much after modification, the word will be replaced with a randomly selected word in the vocabulary. Due to the lack of constructions, this attack sometimes changes the semantics of the original sentence.  
	
	Jin \etal{} \cite{djzjjjtzps2020} presented Textfooler, a black-box attack to fool the bidirectional encoder representations from transformer model (BERT) on text classification. They first identify the important words for the target model and then prioritize to replace them with synonyms until the prediction is altered. The word importance $I_{w_i}$ is calculated as 
	\begin{eqnarray}
	I_{w_i}=\left\{
	\begin{array}{ll}
	F_Y(X)-F_Y(X_{w_i}),if F(X)=F(X_{w_i})=Y \\
	F_Y(X)-F_Y(X_{w_i})+F_{\hat{Y}}(X_{w_i})-F_{\hat{Y}}(X), \\
	if F(X)=Y,F(X_{w_i})=\hat{Y}, and \quad Y\neq \hat{Y}
	\end{array} 
	\right.
	\end{eqnarray}
	where $w_i$ is the $i$-th word in $X$. $F_Y(\cdot)$ represents the prediction score for the $Y$ label.   
	
	Considering the limitations (\textit{i.e.,} out-of-context and unnaturally complex token replacements) of synonym substitution methods, Garg \etal{} \cite{sggRamakrishnan2020} used contextual perturbations from a BERT masked language model to generate adversarial examples. They search for important words similar to Jin \etal{} \cite{djzjjjtzps2020}. Then, words from the pre-trained BERT masked language model are used to replace the important words in the inputs or inserted to adjacent positions.
	
	\subsubsection{Other attacks}
	
	Apart from gradient-based and importance-based attacks, researchers also propose other ways to create adversarial examples such as the genetic algorithm and particle swarm optimization-based search algorithm.
	
	Alzantot \etal{} \cite{maysaebkmskwc2018} proposed a continuous optimization method at the word-level using the genetic algorithm (GA) \cite{ejamcf1994,hm1989}. GA initializes the generation by nearest neighbor replacement. The optimization is computed as
	\begin{equation} \label{nearestneighborreplacement}
	x_{adv}=\mathcal{P}^{g-1}_{\arg\max f(\mathcal{P}^{g-1})_{target}} 
	\end{equation}
	where $\mathcal{P}^{g-1}$ represents the $(g-1)$-th generation, $f$ is the target model, $x_{adv}$ means the best individuals in this generation that can fool $f$ to produce incorrect predictions. If the samples in $x_{adv}$ do not satisfy the requirements, they are selected as the next generation to repeat the previous optimization process, where $\mathcal{P}^{g}={x_{adv}}$. Different from DeepWordBug \cite{jgjlmlsyjq2018}, GA utilizes Euclidean distance to maintain the semantics. 
	
	Zang \etal{} \cite{yzfqchyzlmzqlms2020} proposed a novel attack model, which incorporated the sememe-based word substitution method and particle swarm optimization-based search algorithm. For each word in a given text, the authors use HowNet to find its sememe and then add the same labeled words to the word list. After that, a particle swarm optimization algorithm is applied to search for adversarial examples in a discrete search space composed of all the word lists. This work solves the lack of search space reduction methods and inefficient optimization algorithms, significantly improving the success rate of adversarial attacks. 
	
	\subsection{Sentence-level Attacks}
	
	Compared with char-level and word-level attacks, the sentence-level attack is more flexible. The modified sentence can be inserted at the beginning, middle, or end of the text when the semantics and grammar are correct. To some extent, the sentence-level attack can be seen as a special kind of word-level attack manipulated by adding some ordered words. This kind of attack usually appears in other NLP tasks like natural language inference (NLI) \cite{mgvsyg2018}, neural machine translation (NMT) \cite{jedldd2018}, reading comprehension (RC) \cite{rjpl2017}, and question answering (QA) \cite{ewprsftyjrg2019}. In the text classification, this kind of attack is much less than others. 
	
	Iyyer \etal{} \cite{mijwkglz2018} designed syntactically controlled paraphrase networks (SCPNS) for generating adversarial examples by grammar conversion, which relied on the encoder-decoder architecture of SCPNS. Given a sequence and a corresponding target syntax structure, the authors encode them by a bidirectional LSTM and decode them by LSTM. The decoder is augmented with soft attention over encoded states \cite{dbkcyb2014} and the copy mechanism \cite{aspjlcdm2017}. They then modify the inputs to the decoder to incorporate the target syntax structure for the generation. The syntactically adversarial sentences can not only fool pre-trained models but also improve the robustness of them to syntactic variation. However, the measurement of paraphrase quality and grammaticality requires much human effort. The structure of the sentence has changed, although the semantic difference is small.
	
	\subsection{Multi-level Attacks}
	
	Multi-level attacks incorporate at least two of the three adversarial attacks to create more imperceptible and high success rate adversarial examples. Therefore, unlike a single method, the multi-level attack calculation is more expensive and more complicated.
	
	Liang \etal{}\cite{blhlmspbxlws2018} utilized the FGSM to determine what, where, and how to insert, remove, and modify. They use the natural language watermarking technique \cite{mavrmcchfkdmsn2001} to ensure generated adversarial examples compromise their utilities. In the white-box scenario, they define hot training phrases and hot sample phrases by computing the cost gradients of inputs. The former sheds light on what to insert, and the latter implies where to insert, remove, and modify. In the black-box scenario, the hot training phrases and hot sample phrases are obtained through the fuzzing technique\cite{msagpa2007}. When an input is fed to the target model, they use isometric whitespace to substitute the origin word each time. The difference between the results before and after modification is the deviation of each word. The larger it is, the more significant the corresponding word is to the classification. Hence, hot training phrases are the most frequent words in the set of inputs, which consist of the largest deviation words for each training sample. Hot sample phrases are the words with the largest deviation for every test sample.
	
	Like one pixel attack\cite{jwsdvvsk2017} in the image domain, a similar method named HotFlip was proposed by Ebrahimi \etal{}\cite{jeardldd2018}. HotFlip is a white-box attack in texts, which relies on an atomic flip operation to swap one character with another by gradient computation. Compared with DeepWordBug \cite{jgjlmlsyjq2018}, the adversarial examples from HotFlip are more imperceptible due to the fewer modifications. The flip operation is represented by 
	\begin{equation} \label{eq10}
	\begin{split}
	\vec{v}_{ijb} = &(\vec{0},\ldots;(\vec{0},\ldots(0,0,\ldots,0,-1,0,\ldots,1,0)_j,\\&\ldots,\vec{0})_i;\vec{0},\ldots)
	\end{split}
	\end{equation}
	The formula \eqref{eq10} means that the $j$-th character of $i$-th word in an example is changed from $a$ to $b$, which are both characters at $\emph{a}$-th and $\emph{b}$-th places in the alphabet. -1 and 1 are the corresponding positions for $a$ and $b$, respectively. The alteration from directional derivative along this vector is calculated to find the biggest growth in the loss $\emph{J}(x, y)$. The procedure of calculation is shown in 
	\begin{equation} \label{biggestincrease}
	\max\nabla_{x}J(x, y)^T\cdot\vec{v}_{ijb} = \mathop{\max}_{ijb}\frac{\partial J^{(b)}}{\partial x_{ij}} - \frac{\partial J^{(a)}}{\partial x_{ij}}
	\end{equation}
	where $x_{ij}$ is a one-hot vector, which denotes the $\emph{j}$-th character of $\emph{i}$-th word, $y$ refers to the corresponding label vector, $T$ is a transpose function. Apart from character-level attack, HotFlip could also be used on word-level by different modifications. Although HotFlip performs well, only a few successful adversarial examples are generated with one or two flips under strict constraints, thus it is not suitable for a large-scale experiment.
	
	Li \etal{}\cite{jfsjtdbltw2019} proposed an attack framework TextBugger for generating adversarial examples, which could mislead the deep learning-based text understanding system in both black-box and white-box settings. Similar to DeepWordBug \cite{jgjlmlsyjq2018}, TextBugger also searches for important words to modify. In the white-box scenario, Jacobian matrix $J$ is used to calculate the importance of each word.
	\begin{equation}
	C_{x_i} = J_{F(i,y)} = \frac{\partial F_y(x)}{\partial x_i}
	\end{equation}
	where $F_y(\cdot)$ represents the confidence value of class $y$, $C_{x_i}$ is the important score of $i$-th word in $x$. Then, similar modification strategies like DeepWordBug are used to generate both character-level and word-level adversarial examples. In the black-box scenario, the authors segment documents into sequences, and then they query the target model to filter out sentences with different predicted labels from the original ones. The odd sequences are sorted in an inverse order according to their confidence scores calculated by the removal operation as
	\begin{equation} \label{removingmethod}
	\begin{split}
	C_{x_i} = &F_y\left(x_1,\ldots,x_{i-1},x_i,x_{i+1},\ldots,x_n\right) \\& - F_y\left(x_1,\ldots,x_{i-1},x_{i+1},\ldots,x_n\right)
	\end{split}
	\end{equation}
	The final modification process is the same as that in the white-box setting.
	
	Compared with Deep-fool \cite{blhlmspbxlws2018} and TextBugger \cite{jfsjtdbltw2019}, Vijayaraghavan \etal{} \cite{Prashanthvdb2019} applied reinforcement learning to generate adversarial examples in a black-box setting, following an encoder-decoder framework. They extract character and word information from inputs encoded to produce hidden representations of words. Then, an attention mechanism is applied to the decoder for identifying the most relevant text units that highly affect the predictions. During the decoding step, the perturbation vectors are added to those units, and the creations are optimized using target model predictions. 	
	
	\section{Adversarial Techniques for Other NLP Tasks} \label{adversarialexamplesonothertasks}
	
	In section \ref{adversarialattacksintext}, we have reviewed adversarial attacks for the text classification task. Next, we solve some other puzzles on adversarial texts, such as which adversarial examples can attack other kinds of NLP systems or applications and how they are generated in these cases. 
	
	\subsection{Attack on Reading Comprehension Systems} \label{RCS}
	
	The reading comprehension task means that the machine answers the query after reading a given context and the corresponding query. As a constraint, the answer to the query must be a paragraph (\ie{}, several consecutive words) that can be found in the original context. Attackers usually modify the content through sentence-level attacks and induce the system to produce a different answer to the query.
	
	To explore whether reading comprehension systems are vulnerable to adversarial examples, Jia \etal{}\cite{rjpl2017} inserted sentence-level adversarial perturbations into paragraphs to test the systems without changing the answers or misleading humans. They extract nouns and adjectives in the question and replace them with antonyms. Meanwhile, named entities and numbers are changed by the nearest word in GloVe embedding space\cite{jprscm2014}. The modified question is transformed into a declarative sentence as the adversarial example, which is then concatenated to the end of the original paragraph. This process is called ADDSENT by the authors. Another way ADDANY randomly chooses words of the sentences to craft. Compared with ADDSENT, ADDANY does not consider the grammaticality of sentences, and it needs to query the model several times. This work's core idea is to draw the models' attention to the generated sequences rather than original sequences to produce incorrect answers. 
	
	Currently, there is no good defense method to resist this kind of attack. The analysis of the coherence of contextual semantics may be helpful for detection.
	
	\subsection{Attack on Natural Language Inference Models} \label{NLI}
	
	Natural language inference is mainly to judge the semantic relationship between two sentences (\ie{}, premise and hypothesis) or two words. To some extent, it can be regarded as a classification task to ensure that the model can focus on semantic understanding.
	
	Li \etal{} \cite{llrmqgxyxxpq2020} designed a word-level attack to affect the model inference of entity relationship. The method follows the same two-step pipeline of importance-based attacks, and the importance of each word $I_{w_i}$ is calculated like 
	\begin{eqnarray} \label{nNaturaLanguageInf}
	I_{w_i}=o_y(S)-o_y(S_{\backslash w_i})
	\end{eqnarray} 
	where $o_y(S)$ denotes the logit output by the target model for correct label $y$, $S_{\backslash w_i}$ represents the rest $S$ except for the word $w_i$. Unlike synonyms or similar words substitution in the embedding space \cite{srydkhwc2019,djzjjjtzps2020,maysaebkmskwc2018}, a BERT is used for the word replacement, ensuring the semantic similarity and grammar-correct of generated adversarial examples. Considering the word segmentation of a text, the authors divide the substitution into two parts. If the important words are single after segmentation, they are iteratively replaced by the candidates calculated via BERT. Otherwise, the phrase containing an important word is iteratively replaced by the candidates, which are phrases either. 
	
	Minervini \etal{}\cite{pmsr2018} cast the generation of adversarial examples as an optimization problem and proposed a novel multi-level attack. The authors maximize the proposed inconsistency loss $J_{I}$ to search for substitution sets $S$ (\textit{i.e.}, adversarial examples) by using a language model as 
	\begin{equation}
	\begin{split}
	\mathop{maximize}\limits_{S} J_{I}(S) = &\left[p(S;body)-p(S;head)\right]_{+}, \\&s.t. \log p_{L}(S)\leq\tau
	\end{split}
	\end{equation}
	where $[x]_{+}=\max(0,x)$. $p_{L}(S)$ refers to the probability of the sentences in $S$.
	\begin{itemize}
		\item $\tau$: a threshold on the perplexity of generated sequences
		\item ${{X_{1},\ldots,X_{n}}}$: the set of universally quantified variables in a rule to sequences in S
		\item $S= \lbrace{X_{1}\to s_{1},\ldots,X_{n}\to s_{n}}\rbrace$: a mapping from $\lbrace{X_{1},\ldots,X_{n}}\rbrace$
		\item $p(S; body)$ and $p(S; head)$: probability of the given rule, after replacing $X_{i}$ with the corresponding sentence $S_{i}$
		\item $body$ and $head$: represent the premise and the conclusion of the NLI rules
	\end{itemize}
	However, the generated adversarial examples may keep different semantics from the original because of ignoring the semantic changes between modifications and original words. 
	
	\subsection{Attack on Machine Translation Models} \label{NMT}
	
	Machine translation means that the machine can automatically map one language to another. Attackers slightly modify the content of an input language, resulting in the failure to obtain the expected translation result.
	
	Belinkov \etal{}\cite{ybyb2018} conducted a char-level black-box attack to explore the vulnerability of three different neural machine translation models \cite{xzjbzylc2015,jlkcth2017,rsofkcabbhjhmj2017}. The authors devise adversarial examples depending on natural and synthetic language errors, including typos, misspellings, etc. Although these generations can easily fool three different models, they are also visible due to grammatical problems. Apart from the black-box attack, Ebrahimi \etal{}\cite{jedldd2018} and Cheng \etal{} \cite{ycljwm2019} generated adversarial examples with gradient optimization in the white-box settings. Compared with Belinkov \etal{}\cite{ybyb2018}, Ebrahimi \etal{}\cite{jedldd2018} have demonstrated that char-level adversarial examples in black-box attacks are much weaker than white-box ones in most cases, and the generations in the word-level from Cheng \etal{} \cite{ycljwm2019} are more fluent and better. 
	
	Zou \etal{} \cite{wzsjhjxxydjjc2020} generated adversarial examples in the word-level via a new paradigm based on reinforcement learning. They construct a generator based on the generative adversarial networks (GANs) \cite{goodfee3422622,arlmschin2016} to create adversarial examples as follows. The environment (\textit{i.e.,} discriminator and victim NMT model) receives the tokens of current sentences into the agent to select suitable candidates for replacing the target tokens. The candidates of each token are collected through the victim NMT model within Euclidean distance. After modification, the discriminator receives these modified sentences and returns the agent a surviving feedback signal. The process is repeated until the termination signal is received. Unlike the previous works, this work balances semantic approximation and attack effects through self-supervision, and both have achieved good results. 
	
	\subsection{Attack on Question and Answer Systems}
	
\begin{figure}[t]
		\centering
		\subfigure[attack on RC]{
			\begin{minipage}[t]{0.5\linewidth}
				\centering
				\includegraphics[width=\linewidth]{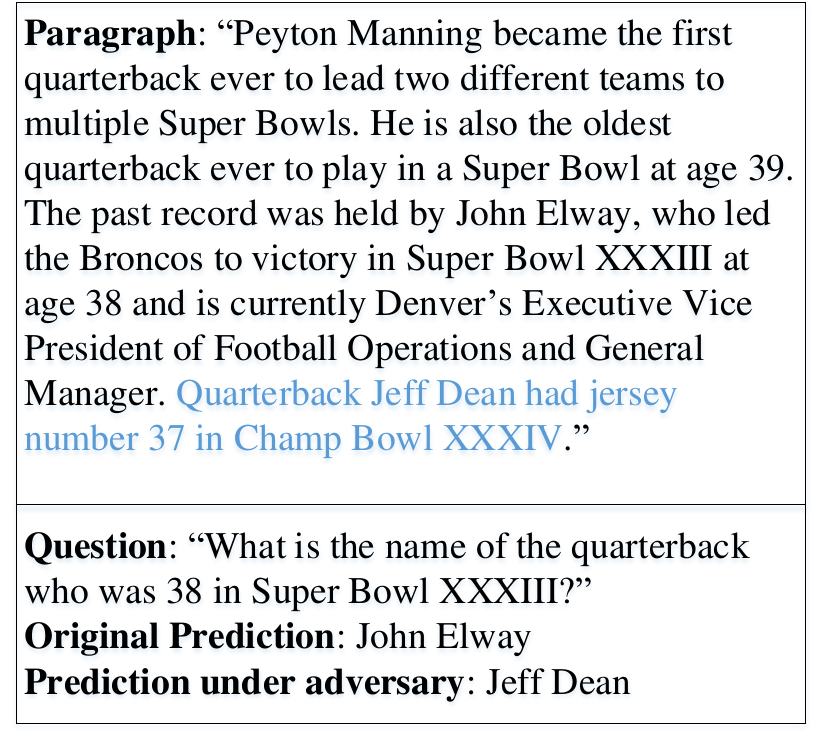}
			\end{minipage}%
		}%
		\subfigure[attack on QA]{
			\begin{minipage}[t]{0.5\linewidth}
				\centering
				\includegraphics[width=\linewidth]{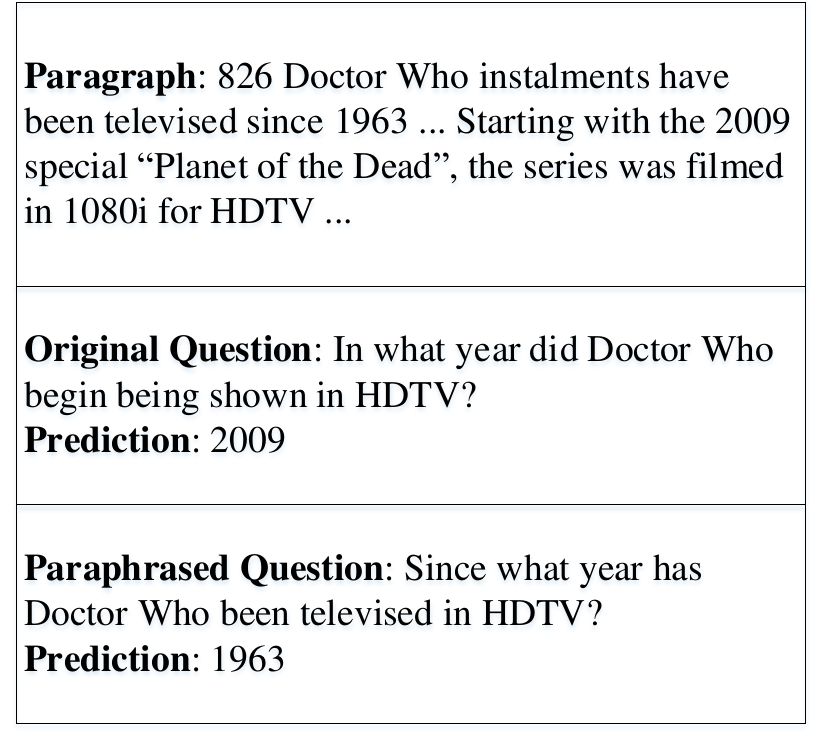}
			\end{minipage}%
		}%
		\centering
		\caption{Adversarial attacks on the machine reading comprehension and question and answer systems. (a) is from Jia \etal{}\cite{rjpl2017}, and (b) is the instance in Gan \etal{} \cite{GanandHweeouNg2019}.}
		\label{mrcandqaaaa}
	\end{figure}

	Inspired by the model's sensitivity to semantically similar questions, Gan \etal{} \cite{GanandHweeouNg2019} generated diverse paraphrased questions in the sentence-level, guiding target models to produce different answers. They obtain all n-grams (up to 6-grams) from the source questions via a language model, and the stopwords in them would be removed. Then, they search the paraphrase database \cite{epprjgbvccb2015} for paraphrases of the remaining. In constrain, the equivalence score between the substitutions and n-grams needs to be greater than 0.25. After paraphrase generation, a pre-trained model \cite{WietingKevin2018} is applied to filter the generated questions with a score greater than 0.95. Compared with adversarial attacks on reading comprehension systems, this work is to modify the question rather than declarative content like Figure \ref{mrcandqaaaa}. Tan \etal{} \cite{stsjmykrs2020} also guided the question and answer model to produce incorrect answers by perturbing the words in the questions. However, the generations are likely to have grammatical errors and easy to be detected.
	
	To deal with the non-differentiable and discrete attributes of texts, Wang \etal{} \cite{bxwhpbpqcswbl2020} proposed a tree-based autoencoder to transfer the discrete text into a continuous representation space for creating adversarial perturbations. They firstly select the adversarial seed (\textit{i.e.,} the input sentence) transferred into a continuous embedding. Next, the optimization similar to C\&W attack \cite{ncdw2017} is conducted upon the embedding to search for perturbations. Finally, the modified embedding is decoded back to adversarial examples with semantic similarity. In the targeted attack for question and answer models, the declarative sentence obtained by reconstructing the target answer and question is treated as an adversarial seed. The generated adversarial examples are inserted into the end of the paragraph like Jia \etal{}\cite{rjpl2017}.  
	
	\subsection{Summary of Adversarial Attacks}  
	
	\begin{table*}[!t]
		\centering
		\caption{Summary of existing adversarial attacks. We mainly show the category, time, work, targeted/non-targeted, black/white, model, data, task, gradient related or not, and project url. For each of them, the papers are sorted by category and time. T/N is short for the targeted/non-targeted attack, W/B is short for the white/black-box attack, C is short for text classification, RC is short for reading comprehension, and TS is short for text summarization.} 
		\label{tabpaper_information}
		\begin{adjustbox}{width=\linewidth,center}
			\begin{tabular}{|c|c|c|c|c|c|c|c|c|c|}
				\hline
				\multirow{3}{*}{Category} & \multirow{3}{*}{Time} & \multirow{3}{*}{Work} & \multirow{3}{*}{\tabincell{c}{Targeted/ \\ Non-targeted}} & \multirow{3}{*}{\tabincell{c}{White/ \\ Black}} & \multirow{3}{*}{Model}  & \multirow{3}{*}{Data} & \multirow{3}{*}{task} & \multirow{3}{*}{Gradient} & \multirow{3}{*}{Project URL} \\
				& & & & & & & & & \tabularnewline 
				& & & & & & & & & \tabularnewline
				\hline 
				\hline
				\multirow{5}{*}{\tabincell{c}{char}} & 2018.1.26 & Gao\cite{jgjlmlsyjq2018} & N & B & LSTM & \tabincell{c}{Enron Spam Dataset \\ IMDB} & C & N & \url{https://github.com/QData/deepWordBug} \tabularnewline
				\cline{2-10}
				& 2018.4.30 & Belinkov\cite{ybyb2018} & N & B & \tabincell{c}{char-CNN \\ Nematus \cite{RicoSennrich2017} \\ char2char \cite{jlkcthsfmann2017}} & \tabincell{c}{WCPC \cite{MaxandGuillaume2010 }, RWSE\footnotemark[9]\cite{TorstenZesch2012} \\ MERLIN \cite{boydeta2014merlin}, MAE \cite{1123413057}} & NMT & N & \url{https://github.com/ybisk/charNMT-noise}	\tabularnewline
				\cline{2-10}
				& 2018.8.20 & Ebrahimi\cite{jedldd2018} & T & binary & char-CNN & TED \cite{CettoloMauro2016} & NMT & Y & \url{https://github.com/jebivid/adversarial-nmt} \tabularnewline
				\cline{2-10}
				& 2019.6.2 & Gil\cite{ygycogjb2019} & N & B & GRU\cite{kcbvmdbyb2014} & Toxic Comment\footnotemark[10] & C & N & \url{https://github.com/orgoro/white-2-black}
				\tabularnewline
				\cline{2-10}
				& 2020.4.7 & Wang\cite{bxwbypxlbl2020} & binary & W & BERT & \tabincell{c}{THUCNews\footnotemark[11] \\ Wechat Finance Dataset} & C & Y & — \tabularnewline
				\hline
				\multirow{21}{*}{\tabincell{c}{word}} & 2016.11.2 & Papernot\cite{nppmas2016} & binary	& W & LSTM & IMDB & C & Y & — \tabularnewline
				\cline{2-10}
				& 2017.4.9 & Samanta\cite{sssm2017} & N & W & CNN & IMDB,twitter\cite{twitterdata} & C & Y	& — \tabularnewline
				\cline{2-10}
				& 2018.7.13 & Sato\cite{msjshsym2018} & N & W & FFNN, LSTM & \tabincell{c}{IMDB,RCV1 \\ Elec\footnotemark[12]\cite{rjtz2015},MR\cite{BoPanLillian} \\ Dbpedia} & C & Y & \url{https://github.com/aonotas/interpretable-adv} \tabularnewline
				\cline{2-10}
				& 2018.7.15 & Mudrakarta\cite{pkmatmskd2018} & T & W & \tabincell{c}{LSTM, NP\cite{anqlmaamda2017} \\ QANet\cite{awyddmtrzkcmnqvl2018}}	& \tabincell{c}{VQA 1.0  \cite{aajlsammclzdbdp2015}, SQuAD \\ WikiTableQuestions\cite{pppliang2015}} & RC,QA & Y & \url{ https://github.com/pramodkaushik/acl18\_results} \tabularnewline
				\cline{2-10}
				& 2018.7.15 & Glockner\cite{mgvsyg2018} & N & B & \tabincell{c}{bi-LSTM\cite{W175308},ESIM \\ DAM\cite{apotddtu2016}} & SNLI & NLI & N & \url{https://github.com/BIU-NLP/Breaking\_NLI} \tabularnewline
				\cline{2-10}
				& 2018.10.31 & Alzantot\cite{maysaebkmskwc2018} & T & B & LSTM,RNN & IMDB,SNLI & C & N & \url{https://github.com/nesl/nlp\_adversarial\_examples} \tabularnewline
				\cline{2-10}
				& 2019.5.12 & Behjati\cite{mbsmmdmsbpfard2019} & binary & W & LSTM & AG’s news,SST & C & Y & — \tabularnewline
				\cline{2-10}
				& 2019.7.28 & Ren\cite{srydkhwc2019} & N & B & char-CNN, LSTM & \tabincell{c}{AG's news,IMDB \\ Yahoo! Answers} & C & N & \url{https://github.com/JHL-HUST/PWWS/} \tabularnewline
				\cline{2-10}
				& 2019.7.28 & Hsieh\cite{ylhmcdcjwwwlhcjh2019} & binary & binary & \tabincell{c}{LSTM,BERT \\ Transformer} & Yelp,MultiNLI,WMT15\footnotemark[2] & C,NMT & N & — \tabularnewline
				\cline{2-10}
				& 2019.7.28 & Cheng\cite{ycljwm2019} & T & W & Transformer\cite{avnsnpjulgkip2017} & LDC corpus, WMT14 & NMT & Y & —  \tabularnewline
				\cline{2-10}
				& 2019.7.28 & Zhang\cite{hzzhznmll2019} & T & binary & bi-LSTM,BiDAF & IMDB,SNLI	& C,NLI	& Y	& \url{https://github.com/LC-John/Metropolis-Hastings-Attacker} \tabularnewline
				\cline{2-10}
				& 2020.2.7	& Jin\cite{djzjjjtzps2020}	& N	& B	& CNN,LSTM,BERT	& \tabincell{c}{AG’s news,IMDB,Fake\footnotemark[13] \\ Yelp,MR,SNLI,MultiNLI} & C,NLI & N & 	\url{https://github.com/jind11/TextFooler} \tabularnewline
				\cline{2-10}
				& 2020.2.7 & Cheng\cite{mcjyhzpyccjh2018} & binary & W & seq2seq & \tabincell{c}{DUC2003, DUC2004 \\ Gigaword, WMT15} & NMT,TS & Y & \url{https://github.com/cmhcbb/Seq2Sick} \tabularnewline
				\cline{2-10}
				& 2020.3 & Yang\cite{pyjccjhjlmij2020} & N & B & CNN,LSTM & IMDB,Yahoo! Answers & C & N	& — \tabularnewline
				\cline{2-10}
				& 2020.7.5 & Zang\cite{yzfqchyzlmzqlms2020}	& T	& B	& LSTM, BERT\cite{jdmwklkt2019}	& IMDB,SST,SNLI	& C,NLI	& N	& \url{https://github.com/thunlp/SememePSO-Attack} \tabularnewline
				\cline{2-10}
				& 2020.7.5	& Zou\cite{wzsjhjxxydjjc2020} & N & W & \tabincell{c}{RNN-Search \cite{dbkcysrgio2015} \\ Transformer} & WMT14 & NMT	& N & — \tabularnewline
				\cline{2-10}
				& 2020.7.5 & Tan\cite{stsjmykrs2020} & N & B & \tabincell{c}{BERT, BiDAF \\ Transformer \\ Seq2Seq \cite{gehring17a}} & SQuAD, WMT14 & QA, NMT & N &	\url{https://github.com/salesforce/morpheus} \tabularnewline
				\cline{2-10}
				& 2020.7.5 & Zheng\cite{xzjzyzcjhmcx2020} & binary & binary & parser\cite{TimothyDozat2017} & English Penn Treebank\cite{mcdmbmcdm2006} & C & Y & 	\url{https://github.com/zjiehang/DPAttack} \tabularnewline
				\cline{2-10}
				& 2020.9.7 & Garg\cite{sggRamakrishnan2020}	& N	& B	& CNN,LSTM,BERT	& \tabincell{c}{Amazon,Yelp,IMDB \\ MR,MPQA\footnotemark[14] \\ SUBJ\cite{bopanglilllee2004},TREC\footnotemark[15] \cite{LiandDan2002}}	& C & N	& —  \tabularnewline
				\cline{2-10}
				& 2020.9.7 & Li\cite{llrmqgxyxxpq2020} & N & B & BERT & \tabincell{c}{Yelp, IMDB, AG’s news \\ SNLI, MultiNLI} & C,NLI & N & \url{https://github.com/LinyangLee/BERT-Attack} \tabularnewline
				\cline{2-10}
				& 2021.2.9 & Maheshwary\cite{rmsmvpudi2021} & N & B & \tabincell{c}{CNN,LSTM,BERT \\ ESIM,InferSent\cite{acdkhslbab2017}} & \tabincell{c}{AG's news, MR, Yelp \\ Yahoo Answers, IMDB \\ SNLI, MultiNLI} & C,NLI & N & — \tabularnewline
				\hline
				\multirow{7}{*}{\tabincell{c}{sentence}} & 2017.9.7 & Jia\cite{rjpl2017} & T & B & \tabincell{c}{Match-LSTM \cite{shwjiang2017} \\ BiDAF \cite{jsakafhh2017}} & SQuAD\cite{prjzklpl2016} & RC & N & \url{https://github.com/robinjia/adversarial-squad} \tabularnewline
				\cline{2-10}
				& 2018.4.30 & Zhao\cite{zzddssss2018} & N & B & LSTM,Google Translate & SNLI & NLI,NMT & N & \url{https://github.com/zhengliz/natural-adversary} \tabularnewline
				\cline{2-10}
				& 2018.7.15 & Ribeiro\cite{mtrsscg2018} & N & B & \tabincell{c}{Visual7W\cite{yzqgmblf2016} \\ fastText\cite{ajegpbjm2017}} & Visual7W data,MR,IMDB & QA,C & N &	\url{https://github.com/marcotcr/sears}  \tabularnewline
				\cline{2-10}
				& 2018.11.16 & Iyyer\cite{mijwkglz2018} & N & B & LSTM & SST,SICK\cite{mmlbmbrrsmrz2014}	& C	& N	& \url{https://github.com/miyyer/scpn} \tabularnewline
				\cline{2-10}
				& 2018.11.16 & Wang\cite{ywangmbsan2018} & T & B & BSAE\cite{mepmnmimgcckl2018} & SQuAD & RC & N & — \tabularnewline
				\cline{2-10}
				& 2019.7 & Wallace\cite{ewprsftyjrg2019} & T & W & RNN,IR\cite{jbgsfpr2017} & Quizbowl questions & QA & N & \url{https://github.com/Eric-Wallace/trickme-interface/} \tabularnewline
				\cline{2-10}
				& 2019.7.28 & Gan\cite{GanandHweeouNg2019} & T & B & \tabincell{c}{BERT \cite{jdmwklkt2019}, DrQA \cite{dcafjwab2017} \\ BiDAF \cite{jsakafhh2017}} & SQuAD & QA & N & — \tabularnewline
				\hline
				\multirow{11}{*}{\tabincell{c}{multi}} & 2018.7.13	& Liang\cite{blhlmspbxlws2018} & T & binary & char-CNN\cite{xzjbzylc2015} & \tabincell{c}{Dbpedia, MR, MPQA \\ Customer review\footnotemark[16]} & C & Y & — \tabularnewline
				\cline{2-10}
				& 2018.7.15 & Ebrahimi\cite{jeardldd2018} & N & W & char-CNN,LSTM \cite{ykyjdsamr2016} & AG’s news & C & Y & \url{https://github.com/AnyiRao/WordAdver} \tabularnewline
				\cline{2-10}
				& 2018.10.31 & Minervini\cite{pmsr2018}	& N	& W	& \tabincell{c}{DAM \cite{apotddju2016}, ESIM \cite{qcxzzlswhjdi2017} \\ bi-LSTM} & SNLI, MultiNLI\cite{awnnsrb2018} &	NLI	& Y	& \url{https://github.com/uclnlp/adversarial-nli} \tabularnewline
				\cline{2-10}
				& 2018.10.31 & Blohm\cite{mbgjesxyntv2018} & N & binary & CNN,LSTM & MovieQA\footnotemark[17] & QA & N & \url{https://github.com/DigitalPhonetics/reading-comprehension} \tabularnewline
				\cline{2-10}
				& 2018.10.31 & Niu\cite{tongniumb2018} & N & B & VHRED\cite{ivasrlrlcjpayb2017},RL\cite{jlwmarmgjgdj2016} & \tabincell{c}{Dialogue Corpus\cite{rlnpisjp2015} \\ CoCoA\cite{hhabmepl2017}} & Dialogue & N & \url{https://github.com/WolfNiu/AdversarialDialogue} \tabularnewline
				\cline{2-10}
				& 2019.2.24	& Li\cite{jfsjtdbltw2019} & N & binary & CNN,LSTM & IMDB,MR & C & Y & — \tabularnewline
				\cline{2-10}
				& 2019.6.2 & Zhang\cite{yzjblhe2019} & N & B & \tabincell{c}{BOW,LSTM,BERT \\ ESIM.DecAtt\cite{apotddju2016} \\ DIIN\cite{yghljzhang2018}} & \tabincell{c}{Quora Question Pairs \\ Wikipedia\footnotemark[18]} & C & N & \url{https://g.co/dataset/paws} \tabularnewline
				\cline{2-10}
				& 2019.9.20	& Vijayaraghavan\cite{Prashanthvdb2019} & N & B & CNN & AG’s news, IMDB & C & N & — \tabularnewline
				\cline{2-10}
				& 2019.11.4	& Wallace\cite{ewsfnkmgss2019} & T & W & \tabincell{c}{Bi-LSTM,ESIM \\ DAM,BiDAF} & SST,SNLI,SQuAD & C,NLI,RC & Y & \url{https://github.com/Eric-Wallace/universal-triggers} \tabularnewline
				\cline{2-10}
				& 2020.9.7 & Wang\cite{bxwhpbpqcswbl2020} & T & W & \tabincell{c}{BERT, Transformer \\ BiDAF} & Yelp, SQuAD & C,QA & Y & \url{https://github.com/AI-secure/T3} \tabularnewline
				\cline{2-10}
				& 2020.12.29 & Li\cite{llysdsxqxjh2020} & N & B & BERT & \tabincell{c}{Sogou,IflyTek \\ Weibo,Law34\footnotemark[19]} & C & N & —  \tabularnewline
				\hline
			\end{tabular}
		\end{adjustbox}
	\end{table*}
	
	Adversarial attack methods have developed rapidly in recent years. Across the four categories (\ie{}, char, word, sentence, and multi-level), high-quality adversarial texts are becoming more difficult to detect by the human eyes. On the other hand, the diversified legitimate and generated texts also promote the development of adversarial attacks and defenses. Therefore, there is still much room for improvement in generating adversarial examples such as transferability and deployment in real-world. 
	
	To demonstrate the generation methods of adversarial texts and their corresponding attributes in detail, we build Table \ref{tabpaper_information} and Table \ref{instance}. Through the two tables, we summarize and analyze the promising trend of the generation method.
	
	\textbf{Possible problems with white-box attacks.} Table \ref{tabpaper_information} summarizes the existing adversarial attacks in texts. We observe that the majority of white-box attacks in Table \ref{tabpaper_information} employ the optimization of gradients for the attack. Gradient-based methods are widely used in the image domain with many variants \cite{akigsb2017,ydfltphsjzxhjl2017}, which can also be applied to texts. However, there are some shortcomings in using the gradients, such as vanishing and exploding gradient problems \cite{ybpspf1994,rptmyb2012} and limitations of the access to target models. In addition, gradient masking \cite{aancdw2018} could incur the gradients useless in some cases, leading to failure in gradient-based methods. 
	
	\footnotetext[9]{\url{https://www.informatik.tu-darmstadt.de/ukp/research\_6/data/index.en.jsp}}
	\footnotetext[10]{\url{https://www.kaggle.com/c/jigsaw-toxic-comment-classification-challenge/}}
	\footnotetext[11]{\url{https://github.com/thunlp/THUCTC}}
	
	\textbf{Black-box attacks in the real world.} Table \ref{tabpaper_information} also shows that the focus of research is on more realistic black-box attacks at present. It is basically difficult for adversaries to obtain target models' full knowledge in the physical world, so they do not know what dataset and model the defenders use. In this case, a black-box attack that can effectively deceive the system deployed in the physical world is expected. However, existing black-box attacks do not satisfy the requirements. We create 1,000 adversarial examples via various black-box attacks \cite{jgjlmlsyjq2018,maysaebkmskwc2018,srydkhwc2019,sggRamakrishnan2020} to evaluate their performance on a physical classification system ParallelDots, but only about 9\% of the samples can successfully fool the system. Although these adversarial examples are not specially designed for ParallelDots, the results also reflect the insufficient transferability of adversarial examples in real applications. On the other hand, if a model's output is the hard label rather than a score (\textit{e.g.,} the output of Figure \ref{isnatncefigure}(a) is 1 rather than 64.30\%), the black-box attacks such as importance-based ones need to be improved and adapted to new situations.
	
	\footnotetext[12]{\url{http://riejohnson.com/cnn_data.html}}
	\footnotetext[13]{\url{https://www.kaggle.com/c/fake-news/data}}
	\footnotetext[14]{\url{http://mpqa.cs.pitt.edu/}}
	\footnotetext[15]{\url{https://cogcomp.seas.upenn.edu/Data/QA/QC/}}
	
	\textbf{Evaluation by human eyes.} Furthermore, outstanding adversarial texts not only achieve a high success rate to fool DNNs, but also need to have good readability, semantic similarity, and imperceptibility. Hence, we can also evaluate generated adversarial examples through instances (in Table \ref{instance}). Modifications on texts are generally divided into char-level, word-level, sentence-level, and multi-level. The char-level operates on the characters, and others modify words or sentences. In Table \ref{instance}, the word-level adversarial examples seem more imperceptible than the char-level ones, although people are robust against misspellings \cite{rg2007}. Nevertheless, some char-level methods also perform very well such as HotFlip \cite{jeardldd2018}. Generally, the more operations there are, the easier it is to be perceived. The more imperceptible the perturbations are, the better the readability and semantic similarity would be.
	
	\begin{table*}[t]
		\centering
		\caption{Instances of some adversarial attacks. The content in parentheses is the modified word or sentence. After modification, the output of the model changes from one category to another. NNR is short for nearest neighbor replacement, ST is short for synonym substitution, GC is short for Grammar conversion.}
		\label{instance}
		\begin{tabular}{|p{1cm}<{\centering}|p{1.7cm}<{\centering}|p{12.5cm}|p{1.1cm}<{\centering}|}
			\hline
			Category & Work & \multicolumn{1}{c|}{Instance} & Operation \\
			\hline
			\multirow{1}{*}{char} & Gao\cite{jgjlmlsyjq2018} & \tabincell{l}{This film has a special place (\textbf{plcae}) in my heart (\textbf{herat}). \quad  \textbf{positive $\rightarrow$ negative}} & swap \\
			\hline
			\multirow{3}{*}{word} & Alzantot\cite{maysaebkmskwc2018} & \tabincell{l}{A runner (\textbf{racer}) wants to head for the finish line. \quad \textbf{86\%Entailment$\rightarrow$43\%Contradiction}} & NNR \\ 
			\cline{2-4}
			& Ren\cite{srydkhwc2019} & \tabincell{l}{seoul allies calm on nuclear (\textbf{atomic}) shock. south korea’s key allies play down a shock admission \\ its scientists experimented to enrich uranium. \quad \textbf{74.25\%Sci/Tech$\rightarrow$86.66\%World}} & ST \\
			\cline{2-4}
			& Garg\cite{sggRamakrishnan2020} & \tabincell{l}{Our server was great (\textbf{enough}) and we had perfect service (\textbf{but}). \quad  \textbf{positive $\rightarrow$ negative}} & insert \\
			\hline
			\multirow{1}{*}{sentence} & Iyyer\cite{mijwkglz2018} & \tabincell{l}{there is no pleasure in watching a child suffer. (\textbf{in watching the child suffer, there is no pleasure.}) \\ \textbf{negative $\rightarrow$ positive}} & GC\\
			\hline
			\multirow{1}{*}{multi} & Liang\cite{blhlmspbxlws2018} & \tabincell{l}{The Old Harbor Reservation Parkways are three \sout{historic} roads in the Old Harbor area of Boston.\\ (\textbf{Some exhibitions of Navy aircrafts were held here.}) They are part of the Boston parkway system \\ designed by Frederick Law Olmsted. They include all of William J. Day Boulevard running from \\ Castle (\textbf{Cast1e}) Island to Kosciuszko Circle along Pleasure Bay and the Old Harbor shore. The part \\ of Columbia Road from its northeastern end at Farragut Road west to Pacuska Circle (formerly  \\ called Preble Circle). \textbf{87.3\%Building$\rightarrow$95.7\%Means of Transportation}} & \tabincell{c}{Insert \\ delete \\ flip} \\
			\bottomrule
		\end{tabular}
	\end{table*}   
	
	\section{Defenses Against Adversarial Attacks in Texts} \label{defense}
	
	The constant arms race between adversarial attacks and defenses invalidates backward methods quickly \cite{xlsjjzjwcw2019}. Advanced adversarial attacks will be developed inadvertently, requiring effective defense methods to fight against the threats. In this section, we describe the proposed efficient methods against adversarial attacks in texts, which can be divided into adversarial example detection and model enhancement. The former directly detect adversarial perturbations, and the latter aims at enhancing the robustness of models.
	
	\footnotetext[16]{\url{https://www.cs.uic.edu/~liub/FBS/sentiment-analysis.html}}
	\footnotetext[17]{\url{http://movieqa.cs.toronto.edu/leaderboard/}}
	\footnotetext[18]{\url{https://dumps.wikimedia.org/}}
	\footnotetext[19]{\url{https://github.com/LinyangLee/CN-TC-datasets}}
	
	\subsection{Detecting Misspellings and Unknown Words}
	
	For defense, it is natural to consider whether it can be directly identified based on the difference between adversarial examples and legitimate texts. In the char-level attacks, the majority of modified words are misspellings due to the operations. Similarly, these words may also be unknown in some word-level attacks (\eg{}, word substitution attack). It naturally comes up with an idea to detect adversarial examples by checking misspellings and unknown words. Unknown words are low-frequency or unseen words in the pre-trained model's vocabulary. Misspellings can also be treated as unknown words, which are out of the vocabulary.
	
	Pruthi \etal{} \cite{dpbdzcl2019} built a word recognition model in front of the downstream classifier to distinguish adversarial examples in the char-level. The recognition model treats misspellings as unknown words. In tackling these words, three feedback mechanisms are applied to deal with them. (1) The model passes it through, regardless of whether it is adversarial. (2) Neutral words like `a' or `the' are used for replacement. (3) The word recognition model is retrained with a larger and less-specialized corpus. This work outperforms the general spelling check and adversarial training. Besides, Li \etal{}\cite{jfsjtdbltw2019} applied a context-aware spelling check service to detect misspellings. However, experimental results show that the detection is effective on char-level modifications and partly useful on word-level attacks. The spelling check method is also not suitable for adversarial examples based on other languages like Chinese \cite{wwrwlwbt2019}.
	
	Furthermore, Zhou \etal{} \cite{yczjyjkwcww2019} introduced three components (\ie{}, discriminator, estimator, and recovery) to the model, which was used to discriminate both char-level and word-level perturbations. The components are trained with the original corpus in the training phase. When new text is fed to the detector, the discriminator classifies each token representation as to the perturbation or not. If a token is marked as adversarial, the estimator generates an approximate embedding vector to replace the token representation for recovery. The detector highly relies on the original corpus, leading to the failure of adversarial examples from other corpora. Meanwhile, the basis for classifying whether a token is perturbed based on neighboring tokens is not clear. 
	
	\subsection{Model Enhancement}
	
	The direct detection of inputs will fail when faced with some word-level and sentence-level attacks without unknown words. Therefore, researchers defend against adversarial attacks by ensuring the security of models. Mainstream model enhancement methods include changing the model architecture or updating the model parameters, such as adversarial training \cite{tmamdig2017}, certified robustness training \cite{jia2019certified}, and functional enhancement \cite{ejrjarpl2020}. 
	
	\subsubsection{Adversarial Training} \label{Adversarialrainin}
	
	In texts, adversarial training and its variants are widely applied to defend against adversarial examples. Researchers mix adversarial examples with the original ones for retraining, improving the models' tolerance to adversarial examples. 
	
	Wang \etal{} \cite{ywangmbsan2018} applied data augmentation to create diverse samples for adversarial training. Compared with the AddSent-trained model \cite{rjpl2017}, abundant data and semantic-relations can overcome model overstability and increase their robustness. Similarly, Wang \etal{} \cite{xswhjkh2019} and Wang \etal{} \cite{wang2020defense} extracted the synonyms for the training data and replaced the original words to construct a larger dataset for retraining. Nevertheless, they are specially designed for the synonym substitution attack and may fail to detect other kinds of adversarial attacks like the char-level ones.
	
	The aforementioned methods \cite{ywangmbsan2018,xswhjkh2019} are traditional adversarial training \cite{ijgjscz2015} that directly constructs the training dataset with all the adversarial examples and legitimate ones. Yet, researchers have demonstrated that traditional adversarial training is weak against iterative attacks and proposed iterative training methods in the image domain \cite{amamlsdtav2018,arXiv190412843S}. Inspired by the iterative methods in images, Liu \etal{} \cite{klxlayjljssslqqs2020} and Liu \etal{} \cite{hlyzzywzlyc2020} created new adversarial examples at each epoch and added them for training. Through the iterative optimization on loss functions, the retrained models are more robust against adversarial examples. 
	
	Unlike the above methods, Xu \etal{} \cite{jjxlzhyqzylxs2019} proposed a novel adversarial training approach LexicalAT based on GANs \cite{goodfee3422622,arlmschin2016}. LexicalAT has two components, generator and classifier. The generator creates adversarial examples with the designed replacement actions, and the classifier returns feedback of this action by calculating the absolute difference in probability between the adversarial examples and the corresponding original texts. Then, the generator maximizes the expectation of the feedback by policy gradient, and the classifier minimizes the loss function until convergence. LexicalAT combines a knowledge base and adversarial learning and improves the robustness of sentiment classification models to a certain degree. Dinan \etal{} \cite{edshbcjw2019} conducted a similar work to construct robust models. Differently, they use crowderworkers instead of the generator in Xu \etal{} \cite{jjxlzhyqzylxs2019}, so their work needs much human effort.
	
	In adversarial training, data diversity is the key factor in determining the robustness of models, which relies on heuristic approximations to the worst-case perturbations. As a result, adversarial training is vulnerable to unknown attacks.
	
	\subsubsection{Functional Improvement}
	
	DNN contains many built-in functions as well as functions that can be added externally. Researchers utilize the specially designed functions to reduce differences in the representation of adversarial examples and legitimate samples in the model, thus eliminating the impact of adversarial perturbations on the model. Here, we introduce existing works in this aspect.
	
	Jones \etal{} \cite{ejrjarpl2020} constructed a robust encoding function to map inputs to a smaller and discrete encoding space. They construct a vocabulary containing the most frequent words in texts. Through clustering words in the dictionary, the words in a cluster share the same encoding. These encodings are the models' training data, \ie{}, $f_{\alpha}=g(\alpha (x))$. $f_{\alpha}$ is the classifier, and $g(\cdot)$ receives the outputs from the encoding function $\alpha$. For an adversarial example, the perturbations and the corresponding original words will be in the same cluster (stability), and the non-perturbations are not affected (fidelity). However, the performance is restricted by the size of the vocabulary. Besides, the trade-off between stability and fidelity also needs more detailed analysis.
	
	Li \etal{}\cite{ahlas2019} incorporated the external knowledge to the multi-head attention \cite{avnsnpjulgkip2017} for enhancing the robustness of NLI systems. The model can search for external knowledge when conducting NLP tasks, helping the model to explore beyond the data distribution of specific tasks. Experimental results show a significant improvement in defending against adversarial examples when the knowledge is added to the cross-encoder in their models. Although the method does not need extra parameters and is suitable for any model with attention units, the quality and size of external knowledge limit the performance of the method. 
	
	\subsubsection{Certification}
	
	The detection of unknown words and adversarial training partially mitigate the threats of adversarial examples, but they are unlikely to find worst-case adversaries due to the complexity of the search space arising from discrete text perturbations \cite{pswrsjwcddysgkdpk2019}. Hence, researchers have proposed certification robustness training to search for a boundary. In some conditions, a model is guaranteed to be robust against an attack, \ie{}, it can not cross the boundary, no matter how adversaries create adversarial examples. 
	
	Jia \etal{}\cite{jia2019certified} presented certified robustness training by optimizing the interval bound propagation (IBP) upper bound \cite{kdsgrsrabojupk2018}, which could limit the loss of worst-case perturbations and compress the living space of adversarial examples. This method is provably robust to the attacks with word substitution on IMDB and SNLI. Similarly, Huang \etal{} \cite{pswrsjwcddysgkdpk2019} also applied IBP for certified robustness training. However, IBP only works to certify DNNs with continuous inputs, so it is not applicable to other models, such as the char-level one \cite{xzjbzylc2015}. Differently, Ko \etal{} \cite{cykzyllwldnwdl2019} proposed a gradient-based approach to find the minimum distortion of neural networks or lower bounds for robustness quantification. It is suitable for various models and has no restrictions like IBP, but it is inefficient and poses a computational challenge for Transformer verification due to the self-attention mechanism.
	
	Considering that the current certification methods only deal with naive DNNs, Shi \etal{} \cite{zxshzkwmhcjh2020} proposed a novel method to certify the robustness of more complex Transformers. A Transformer model is decomposed into a number of sub-layers. Each sub-layer contains multiple positions, and each position consists of multiple neurons. The global lower and upper bounds of each neuron (\textit{w.r.t.}, the input within the perturbation space) are calculated to efficiently obtain a safety guarantee by reducing the distance between bounds. Compared with IBP-based methods \cite{jia2019certified,pswrsjwcddysgkdpk2019}, the certified robustness bounds in this work are much tighter, and they also identify word importance as the same as importance-based methods.
	
	Ye \etal{} \cite{mycygql2020} designed a structure-free certified defense method that can guarantee the robustness of any pre-trained model. They construct a smoothed classifier $g^{RS}$ by introducing random substitutions from a synonym set, where $R$ represents the perturbed words and $S$ refers to the corresponding substitutions. The certification of the newly constructed model is defined as
	\begin{eqnarray}
	\Delta x \overset{def}{=} \min_{X'\in S_X}g^{RS}(X',y)-\max_{X'\in S_X}g^{RS}(X',c)  >0
	\end{eqnarray}
	where $X'$ represents the modified sentences in synonym set $S_X$. $c$ is any label except the true label $y$. If the lower bound is larger than the upper bound (\textit{i.e.,} $\Delta x$), the smoothed classifier is certified robust.
	
	However, this kind of method is largely affected by models, testing data, and optimization methods. It is not general, as the detection of unknown words and adversarial training.
	
	\begin{table*}[t]
		\scriptsize
		\centering
		\caption{Summary information of defense methods against adversarial examples. We mainly show the category, time, work, model, data, attack, and project url.}
		\label{defenseinformation}
		\begin{adjustbox}{width=\linewidth,center}
			\begin{tabular}{|c|c|c|c|c|c|c|}
				\hline 
				\multirow{3}{*}{Category} & \multirow{3}{*}{Time} & \multirow{3}{*}{Work} & \multirow{3}{*}{Model}  &  \multirow{3}{*}{Attack} & \multirow{3}{*}{NLP Task} & \multirow{3}{*}{Project URL} \\
				& & & & & &  \tabularnewline
				& & & & & &  \tabularnewline
				\hline 
				\hline
				\multirow{3}{*}{Detection} & 2019.2.24 & Li\cite{jfsjtdbltw2019} & Microsoft Azure & char-level & C & — \tabularnewline
				\cline{2-7}
				& 2019.7.28 & Pruthi\cite{dpbdzcl2019} & bi-LSTM, BERT & word, char-level & C & \url{https://github.com/danishpruthi/adversarial-misspellings} \tabularnewline
				\cline{2-7}
				& 2019.11.4 & Zhou\cite{yczjyjkwcww2019} & BERT & word,char-level & C & \url{https://github.com/joey1993/bert-defender} \tabularnewline
				\hline
				\multirow{7}{*}{\tabincell{c}{Adversarial \\ training}} & 2018.11.16 & Wang\cite{ywangmbsan2018} & BSAE & sentence-level & RC & — \tabularnewline
				\cline{2-7}
				& 2019.9.15 & Wang\cite{xswhjkh2019} & CNN,LSTM & word-level & C & — \tabularnewline
				\cline{2-7}
				& 2019.11.4 & Xu\cite{jjxlzhyqzylxs2019} & CNN,LSTM,BERT & word-level & C & \url{https://github.com/lancopku/LexicalAT} \tabularnewline
				\cline{2-7}
				& 2019.11.4 & Dinan\cite{edshbcjw2019} & BERT & char,word-level & dialogue & —  \tabularnewline
				\cline{2-7}
				& 2020.2.7 & Liu\cite{klxlayjljssslqqs2020} & \tabincell{c}{QANet, BERT \\ ERNIE2.0 \cite{ysswylsfhthwhw2020}} & sentence-level & RC & —  \tabularnewline
				\cline{2-7}
				& 2020.2.7 & Liu\cite{hlyzzywzlyc2020} & char-CNN. LSTM & char, word-level & C & —  \tabularnewline
				\cline{2-7}
				& 2020.8.28 & Wang\cite{wang2020defense} & CNN,LSTM & word-level & C & \url{https://github.com/Raibows/RSE-Adversarial-Defense} \tabularnewline
				\hline
				\multirow{2}{*}{\tabincell{c}{Functional \\ Improvement}} & 2019.8.31 & Li\cite{ahlas2019} & \tabincell{c}{DAM \cite{apotddju2016}, BERT \\ Transformer} & word-level & NLI &  —  \tabularnewline
				\cline{2-7}
				& 2020.7.5 & Jones\cite{ejrjarpl2020} & BERT & char-level & C,NLI & —  \tabularnewline
				\hline
				\multirow{5}{*}{Certification} & 2019.6.9 & Ko\cite{cykzyllwldnwdl2019} & LSTM & char,word-level & C & \url{https://github.com/ZhaoyangLyu/POPQORN} \tabularnewline
				\cline{2-7}
				& 2019.7.28 & Huang\cite{pswrsjwcddysgkdpk2019} & CNN & char,word-level & C & \url{https://github.com/deepmind/interval-bound-propagation/tree/master/examples/language/} \tabularnewline
				\cline{2-7}
				& 2019.11.4 & Jia\cite{jia2019certified} & BOW,CNN,LSTM & word-level & C,NLI & \url{https://github.com/robinjia/certified-word-sub} \tabularnewline
				\cline{2-7}
				& 2020.4.30 & Shi\cite{zxshzkwmhcjh2020} & Transformer &  word-level &  C  & \url{https://github.com/shizhouxing/Robustness-Verification-for-Transformers} \tabularnewline
				\cline{2-7}
				& 2020.7.5 & Ye\cite{mycygql2020} & CNN, BERT & word-level & C & \url{https://github.com/lushleaf/Structure-free-certified-NLP} \tabularnewline
				\cline{2-7}
				& 2020.8.12 & Li\cite{jfltdsljrzqlmytw2020} & \tabincell{c}{TextCNN \cite{YoonKim2014} \\ bi-LSTM} & char,word-level & C & —  \tabularnewline
				\hline
			\end{tabular}
		\end{adjustbox}
	\end{table*}
	
	\subsection{Theoretical Analysis} 
	
	The aforementioned methods shown in Table \ref{tabpaper_information} and Table \ref{defenseinformation} are actual ways for adversarial attacks and defenses, but none of them explain theoretically why NLP models give different predictions. However, analyzing and explaining models' abnormal behavior is the fundamental way to carry out or solve adversarial attacks, which is lacking at present. 
	
	At present, the related model analysis works in NLP take legitimate data as inputs and observe the behavior of DNNs. According to the objects, we divide analysis methods \cite{ybjglass2019} into two categories: external input and model's internal structure. These works have confirmed the theoretical correctness of some existing methods. They also help us have a better understanding of DNNs and then propose stronger attacks and defenses. 
	
	\textbf{External input.} Studies have demonstrated that the changes of external inputs (\eg{}, input composition \cite{lagmkrmws2017} or representation \cite{yaekyboflyg2017,psbjoa2017}) will affect the outputs of models. For example, Arras \etal{} \cite{lagmkrmws2017} extended the layer-wise relevance propagation (LRP) method to LSTM, producing reliable explanations of which words were responsible for attributing sentiment in individual texts. Gupta \etal{} \cite{pghschutze2018} proposed layer wise semantic accumulation (LISA) method to explain how to build semantics for a recurrent neural network (RNN) and how the saliency patterns act in the decision. During these findings, the authors analyze the sensitiveness of RNNs about different inputs to check the increase or decrease in prediction. The two works prove the theoretical correctness of these importance-based attacks, such as DeepWordBug \cite{jgjlmlsyjq2018}, PWWS \cite{srydkhwc2019}, and Textfooler \cite{djzjjjtzps2020}.
	
	\textbf{Internal structure.} Exploring the performance of the internal units of the model is a more effective analysis method. Aubakirova \etal{} \cite{Malikaamb2016} presented activation clustering to track the maximally activated neurons. Similarly, Dalvi \etal{} \cite{fdndhsybdabjg2019} studied individual neurons capturing certain properties that are deemed important for the task. Through this way, they can increase model transparency and uncover the importance of the individual parameters, helping understand the inner workings of DNNs. Researchers have realized the defense methods by operating the neurons in the image domain \cite{ssnrppkr2020,cz9286885}. Whether it is feasible in texts is worth exploring. 
	
	Jacovi \etal{} \cite{ajossyg2018} presented an analysis into the inner workings of CNNs for processing text. They have demonstrated that the filters capture semantic classes of ngrams, and max-pooling separates the ngrams related to the final classification from the others. By inserting several ngrams, the filters will produce the results beyond extraction, leading to misclassification. Wallace \etal{} \cite{ewjtjwssmgss2019} applied HotFlip \cite{jeardldd2018} to the AllenNLP for interpreting models' weaknesses. However, they simply analyze the realization of different models and do not go deep into the network's internal behavior, contributing to the implementation of defense methods.
	
	Indeed, researchers can combine adversarial examples with existing model analysis methods to explore and analyze models' behavior, such as which layer of the model changes the prediction and differences in propagation path (\ie{}, composed of activated neurons) between adversarial examples and legitimate inputs. The combination can inspire us to come up with more effective ways to eliminate the vulnerability of models. 
	
	\section{Adversarial examples in Chinese-based models} \label{Chinesebasedmodels}
	
	English and Chinese are the two most popular languages in the world. However, DNN's processing of two language inputs is different, resulting in the abnormal performance of adversarial examples in the two languages. Next, we introduce the works in adversarial attacks and defenses targeting the Chinese.
	
	\subsection{Attack}
	
	Adversarial attacks in Chinese-based models are different from those in English due to the text attributes. First, Chinese texts need segmentations before feeding to the models. Second, each token after segmentation is a signal character, word, or phrase. The operations, such as swapping in the char-level and simple substitution of a phrase in the word-level, are not suitable for better generations in Chinese. To deal with these challenges, researchers investigate and design new ways to generate adversarial Chinese texts. 
	
	Wang \etal{} \cite{bxwbypxlbl2020} proposed a Chinese char-level attack against BERT, which could map the discrete text into a high-dimensional embedding space. Due to the mapping capability of BERT, attackers can search the high-dimensional embedding space for modification. The perturbed embeddings are mapped back to the characters with the closest semantics. In the targeted attack scenario, the optimization of the objective function $g(\cdot)$ following C\&W attack \cite{ncdw2018} is defined as 
	\begin{eqnarray}
	g(x')=\max \left[\max\left\{f(x')_i:i\neq t\right\}-f(x')_t,-\kappa\right]
	\end{eqnarray}
	where $f(x')_i$ is the $i$-th element in a logit vector from BERT model $f$. $\kappa$ encourages the optimization to find perturbed character $x'$ classified as class $t$ with high probability. 
	
	In the non-targeted attack scenario, $g(\cdot)$ is slightly different from the targeted attack scenario.
	\begin{eqnarray}
	g(x')=\max \left[f(x')_t-\max\left\{f(x')_i:i\neq t\right\},-\kappa\right]
	\end{eqnarray}
	Here, $t$ is the original class of the input. However, the embedding of $x'$ is closest to the original character $x$, but their semantics may be different, and sometimes $x'$ is unnatural to Chinese readers. 
	
	Li \etal{} \cite{llysdsxqxjh2020} followed the importance-based methods \cite{djzjjjtzps2020,llrmqgxyxxpq2020} to quantify the importance of each segmentation replaced by the pieces in pre-constructed vocabulary. The pieces are similar to the segmentation, which can be a signal character, word, or phrase. The generations in this work are more natural and semantically similar than Wang \etal{} \cite{bxwbypxlbl2020}. We treat the phrases in pieces as special sentences that are shorter than normal, so that the attack can be seen as a multi-level one. 
	
\begin{figure}[t]
		\centering
		\subfigure[Wang et al.]{
			\begin{minipage}[t]{0.5\linewidth}
				\centering
				\includegraphics[width=\linewidth]{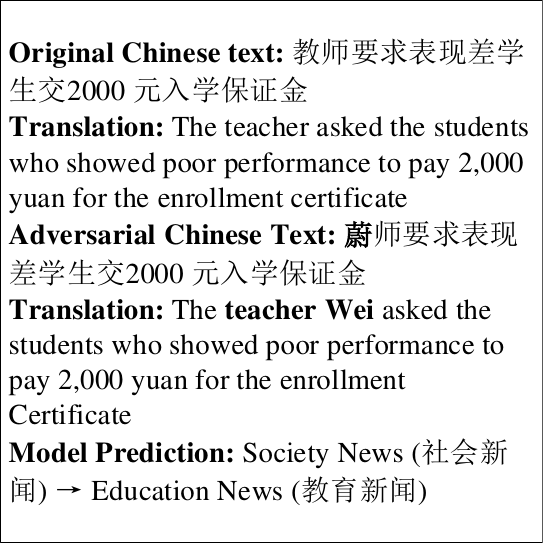}
			\end{minipage}%
		}%
		\subfigure[Li et al.]{
			\begin{minipage}[t]{0.5\linewidth}
				\centering
				\includegraphics[width=\linewidth]{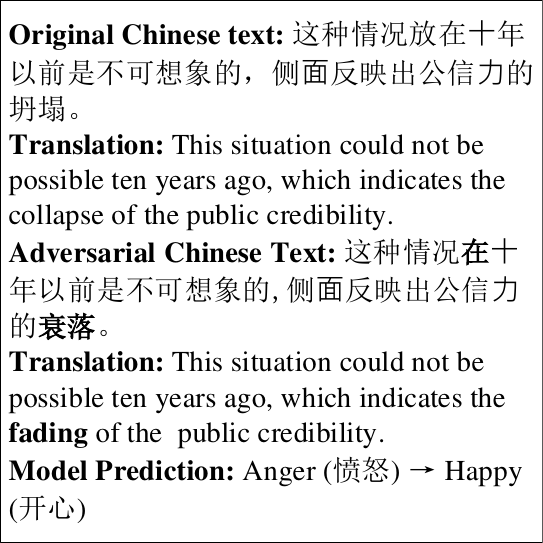}
			\end{minipage}%
		}%
		\centering
		\caption{Adversarial examples in Chinese. (a) is Wang \etal{} \cite{bxwbypxlbl2020}, and (b) is the instance in Li \etal{} \cite{llysdsxqxjh2020}.}
		\label{chinesebasedsamples}
	\end{figure}

	Adversarial examples in Chinese are shown in Figure \ref{chinesebasedsamples}. When the Chinese text is slightly modified, the prediction of the model is converted from one to another. However, the meaning of translations changes more obviously. If we feed the Chinese text and its translation into Chinese-based and English-based classifiers respectively, the consistency of two models' outputs is worth exploring as a basis for judging adversarial examples.
	
	
	\subsection{Defense} 
	
	In tackling the adversarial attacks, countermeasures are urgently needed to handle the threats. Nevertheless, there are few defense methods against adversarial texts, let alone against Chinese texts. The mainstream methods to enhance the robustness of English-based models are adversarial training and spelling check, but they are sometimes not suitable to Chinese-based models. The reasons are shown below.
	\begin{itemize}
		\item The typos in Chinese refer to additional/missing words, wrong word sequence, homophonic/similar words, and semantic errors. Spelling check for English can not be applied to Chinese.
		\item Adversarial training requires a lot of data to achieve a good result and is always sensitive to unknown attacks. The rich meaning and diverse composition of Chinese make it easy to generate adversarial examples. Simply modifying the words can change the semantics. Hence, it is difficult to defend against such diverse Chinese-based adversarial examples.
	\end{itemize}
	
	To bridge this striking gap, Li \etal{} \cite{jfltdsljrzqlmytw2020} proposed TEXTSHIELD, a defense method specially designed for Chinese-based text classification models. The main components of TEXTSHIELD are the NMT model trained with adversarial–benign text pairs and the feature extraction framework. When a Chinese text is fed to TEXTSHIELD, it is first translated into English and then back into Chinese. Finally, the Chinese-based models extract the semantic, glyph, and phonetic-level features of corrected Chinese texts and fuse them for classification. The translation avoids the interference of perturbations from raw texts, and multi-modal embedding features provide more valuable information for classification. However, the performance of TEXTSHIELD highly relies on the two pre-trained models, NMT and the Chinese text classification model. TEXTSHIELD will fail when adversarial examples can fool both of those two models.
	
	\section{Discussions} \label{discussion}
	
	In the previous sections, detailed descriptions of adversarial attacks and defenses in texts enable readers to have a faster and better understanding of this aspect. Next, we present more general observations and shed some light on further work in this area.
	
	\subsection{Generation of Adversarial Examples}
	
	We have reviewed over 40 published or pre-printed papers on the topic of adversarial example generation. In reviewing these attacks, we have some interesting findings, including challenges, which may shed new light on designing more powerful attacks.
	
	\textbf{Limitation of char-level attacks.} Compared with word-level and sentence-level attacks, the char-level attacks are more obvious to human eyes and easier to be detected by some spelling-check tools. Besides, it is difficult to generate an outstanding sample by only modifying one or two characters. In most cases, people have to increase the number of modified characters to generate adversarial examples, resulting in reduced imperceptibility and readability. 
	
	\textbf{The failure of transferability in reality.} Currently, the majority of studies on adversarial texts are about theoretical models and rarely related to practical applications. We have used the adversarial examples presented in recent works\cite{jfsjtdbltw2019,nppmas2016,jgjlmlsyjq2018,maysaebkmskwc2018,jeardldd2018,msjshsym2018,blhlmspbxlws2018,sssm2017} to attack ParallelDots like Figure \ref{isnatncefigure}, but most of the adversarial examples are ineffective and can be correctly classified. Only a few samples successfully fool this system, which means that the transferability of these adversarial examples is bad. For the physical NLP systems, we can not obtain any knowledge of them, and the query may be limited sometimes. Hence, transferability is the main choice for attacking these physical applications, which is the key factor for practical attacks. 
	
	\textbf{Lacking general methods.} There are no well-performed adversarial perturbations in texts that can fool any DNN-based model (so-called universal adversarial perturbations). Although Wallace \etal{}\cite{ewsfnkmgss2019} find input-agnostic sequences that can trigger specific classifications to generate universal adversarial examples, these sequences impact the readability of inputs, and the generated samples are offensive in nature. 
	
	\textbf{Lacking better evaluation methods.} Most of the studies evaluate their performances of adversarial attacks by using success rate or accuracy. Only a few works\cite{jgjlmlsyjq2018},\cite{jfsjtdbltw2019} employ speed, scale, and efficiency into consideration, although they only list the attacks' time. Whether there is a relationship among the scale of the dataset, time consumed, and success rate of adversarial attacks is still unknown. If there exists such a relationship, the trade-off of these three aspects may be a research point in future work, like the related study\cite{hzyatfla2019} of speed in adversarial examples. Besides, the experimental results on different datasets are various when the attack method is the same. Whether the type and amount of data may affect adversarial attacks is worth pondering.

	\subsection{Defense Methods Against Adversarial Attacks}
	
	We have reviewed nearly 20 published or pre-printed papers on the topic of defense methods against adversarial attacks. In reviewing these methods, we have some interesting findings, including challenges, which may shed new light on designing more robust models.
	
	\textbf{Application of adversarial examples.} In order to ensure the safety of the model, we can employ adversarial samples to expose its vulnerabilities for further improvements actively. For example, Blohm \etal{} \cite{mbgjesxyntv2018} generated adversarial examples to discover the limitations of their machine reading comprehension model. In different scenarios\cite{xyyphqzxl2019}, their model is robust against meaning-preserving lexical substitutions but fails in importance-based attacks. Fortunately, some other attributions (\eg{}, answer by elimination via ranking plausibility\cite{jehkjh2005}) can be added to improve the model's performance. Cheng \etal{} \cite{mcjyhzpyccjh2018} proposed a projected gradient method to verify the robustness of seq2seq models. They find that seq2seq models are more robust to adversarial attacks than CNN-based classifiers. Through the various adversarial examples, we can know which features, functions, or models can better resist these attacks and guide us where to start and how to improve.
	
	\textbf{Lacking beachmarks.} Various methods have been proposed to study adversarial attacks and defenses in texts, but there is no benchmark. Researchers use different datasets (in Section \ref{datasets}) in their works, making it difficult to compare these methods' advantages and disadvantages. Meanwhile, it also affects the selection of metrics. There is no exact statement about which metric measure is better in a situation and why it is more useful than others. Some comparisons have been made in Textbugger \cite{jfsjtdbltw2019} with several metrics. The best one in this work may be only suitable for it, but ineffective in other works. 
	
	\textbf{Generalization abilities of detectors.} Tackling the unknown adversarial attacks is one of the main challenges for defense. In the past four years, researchers have been working towards this goal to design a general method. However, none of the existing works meet this need. We think that future work can focus more on designing a general defense to a single NLP task and then extending it to other NLP tasks. 
	
	\textbf{A Platform for research.} In terms of a quick start in this aspect, it is necessary to establish an open-source toolbox (\eg{}, AdvBox\cite{advbox} and cleverhans\cite{papernot2016cleverhans} in the image domain) for the research on adversarial texts. The toolboxes in the image domain integrate existing representative methods of generating adversarial images. People can easily do some further studies by them, which reduce time consumption for repetition and promote the development in this field. Compared with those in the image domain, the visual analytics framework proposed by Laughlin \etal{}\cite{blcckskek2019} lacks diverse attack and defense methods. TextAttack \cite{morris2020textattack} contains some representative attacks, including char-level, word-level, and sentence-level. If more attack and defense methods can be incorporated into it, the toolbox will become more powerful.
	
	\subsection{Further Work}
	
	In future work, studies on adversarial examples can start from the following aspects: As an attacker, it is worth designing universal perturbations as they work in the image domain \cite{smmdafofpf2017}. Any text with universal perturbations can induce a model to produce the incorrect output. Moreover, more wonderful universal perturbations can fool multi-model or any model on any text. On the other hand, enhancing the transferability is meaningful in more practical black-box attacks, and the combination of optimization-based and transferability-based methods is another viable way like the work in \cite{fsjcdeyt2020}. On the contrary, defenders prefer to revamp this vulnerability in DNNs completely, but it is no less difficult than redesigning a network. Both of them are long and arduous tasks with the common efforts of many people. At the moment, defenders can draw on methods from the image area to text for improving the robustness of DNNs, \textit{e.g.}, adversarial training\cite{amamlsdtav2018}, adding extra layer\cite{fmpbjmfmg2018}, optimizing cross-entropy function\cite{skmkq2019,knoossskr2019}, or weakening the transferability of adversarial examples.
	
	Alongside, the combination of deepfake and adversarial examples (also called AdvDeepfakes) is a worthy research direction. Deepfake \cite{wangrun2020} refers to a technique to naturally synthesize human imperceptible fake images and editing images via artificial intelligence (AI), especially through GANs. In response to this emerging challenge, researchers have constructed various deepfake detectors \cite{davnjyie2018,sywowrzaoaae2020}, but they fail to detect AdvDeepfakes \cite{pnsshmjfkjm2020,GandhiandShomik2020,NicholasFarid2020,nrsabss2020,pnbdjbccf2020} where attackers add adversarial perturbations to deepfake images. Inspired by these works, whether the fake text detectors \cite{wzdtzxrwndmzjwjy2020,zrharhbyfarfcy2019,ywwyfmjxbzqdjg2020} are robust against AdvDeepfakes needs further exploration. On the other hand, it also encourages researchers to build more robust detectors.
	
	\section{Conclusion} \label{conclusion}
	
	This paper presents a comprehensive survey about adversarial attacks and defenses on DNNs in texts. Although DNNs have a high performance on a wide variety of NLP tasks, they are inherently vulnerable to adversarial examples. Hence, people pay great attention to the security problem caused by adversarial examples. We integrate the existing adversarial attacks and defenses focusing on recent works in texts. The threats of adversarial attacks are real, but defense methods have fallen far behind. Most existing works have their limitations, like application scenes and constraint conditions. More attention should be paid to the problem caused by adversarial examples, which remains an open issue for designing considerably robust models against adversarial attacks.

	\section*{Acknowledgments}
	This work was partly supported by the National Natural Science Foundation of China under No. 61876134, U1536204 and U1836112.
	
	\bibliographystyle{IEEEtran}
	\bibliography{reference}

\end{document}